\pdfoutput=1

\documentclass[a4paper,fleqn]{cas-sc}
\usepackage[authoryear]{natbib}
\usepackage{caption}
\usepackage{subcaption}
\usepackage{hyperref}

\begin{document}
\let\WriteBookmarks\relax
\def\floatpagepagefraction{1}
\def\textpagefraction{.001}
\shorttitle{Post-interactive Multimodal Trajectory Prediction for Autonomous Driving}
\shortauthors{Ziyi Huang et~al.}

\title [mode = title]{Post-interactive Multimodal Trajectory Prediction for Autonomous Driving}                      

\tnotetext[1]{This work was supported by the National Natural Science Foundation of China with 52302493 and 52372411, Hunan Provincial Natural Science Foundation of China with 2023JJ10008, and State Key Laboratory of Advanced Design and Manufacturing Technology for Vehicle with 72275004.}

\author[1]{Ziyi Huang}
\ead{hzy03@hnu.edu.cn}

\credit{Data curation, Formal analysis, Investigation, Methodology, Software, Validation, Visualization, Writing - original draft}

\affiliation[1]{organization={College of Computer Science and Electronic Engineering, Hunan University}, city={Changsha}, postcode={ 410082},country={China}}

\author[2]{Yang Li}
\cormark[1]
\ead{lyxc56@gmail.com}
\credit{Investigation, Writing - review \& editing, Supervision, Resources, Funding acquisition, Project administration, Conceptualization}
\author[2]{Dushuai Li}
\ead{lidushuai@163.com}
\credit{Investigation, Formal analysis, Writing - review \& editing}

\affiliation[2]{organization={College of Mechanical and Vehicle Engineering, Hunan University}, city={Changsha}, postcode={ 410082},country={China}}

\author[3]{Yao Mu}
\ead{muyao@connect.hku.hk}
\credit{Methodology, Formal analysis, Writing - review \& editing}

\affiliation[3]{organization={Department of Computer Science, The  University of Hong Kong}, city={Hong Kong}, postcode={ 999077},country={China}}

\author[2]{Hongmao Qin}
\ead{qinhongmao@hnu.edu.cn}
\credit{Writing - review \& editing, Supervision, Resources}

\author[4]{Nan Zheng}
\cormark[1]
\ead{Nan.Zheng@monash.edu}
\credit{Conceptualization, Writing - review \& editing, Supervision, Resources}

\affiliation[4]{organization={Department of Civil Engineering, Monash University},
                addressline={Wellington Rd}, 
                city={Clayton VIC},
                postcode={3800}, 
                country={Australia}}

\cortext[cor1]{Corresponding author}


\begin{abstract}
Modeling the interactions among agents for trajectory prediction of autonomous driving has been challenging due to the inherent uncertainty in agents' behavior. The interactions involved in the predicted trajectories of agents, also called post-interactions, have rarely been considered in trajectory prediction models. To this end, we propose a coarse-to-fine Transformer for multimodal trajectory prediction, \textit{i.e.}, Pioformer, which explicitly extracts the post-interaction features to enhance the prediction accuracy. Specifically, we first build a Coarse Trajectory Network to generate coarse trajectories based on the observed trajectories and lane segments, in which the low-order interaction features are extracted with the graph neural networks. Next, we build a hypergraph neural network-based Trajectory Proposal Network to generate trajectory proposals, where the high-order interaction features are learned by the hypergraphs. Finally, the trajectory proposals are sent to the Proposal Refinement Network for further refinement. The observed trajectories and trajectory proposals are concatenated together as the inputs of the Proposal Refinement Network, in which the post-interaction features are learned by combining the previous interaction features and trajectory consistency features. Moreover, we propose a three-stage training scheme to facilitate the learning process. Extensive experiments on the Argoverse 1 dataset demonstrate the superiority of our method. Compared with the baseline HiVT-64, our model has reduced the prediction errors by 4.4\%, 8.4\%, 14.4\%, 5.7\% regarding metrics minADE$_6$, minFDE$_6$, MR$_6$, and brier-minFDE$_6$, respectively. 
\end{abstract}


\maketitle

\section{Introduction}

Trajectory prediction for autonomous vehicles (AVs) aims to forecast the future paths or motions of the traffic participants based on past trajectories and environmental clues such as lanes and obstacles \citep{TIV_survey, Pedestrian}, which is essential to the safe and efficient decision and planning of AVs, especially in dynamic urban conditions \citep{PIH, planner}. With accurate trajectory predictions, the vehicle can make proactive decisions that can maximize traffic efficiency while satisfying safety constraints. Trajectory prediction methods can be divided into deterministic prediction and multimodal prediction \citep{huang2023multimodal}, in which the former provides only one prediction for each agent, and the latter can generate multiple predictions each time. To handle the uncertainty and the multimodality involved in driving behavior, this study focuses on multimodal trajectory prediction that can generate multiple traffic rule-compliant predictions based on observed data and context information.     

\begin{figure}[t]
    \centering	   
    \includegraphics[width=0.75\linewidth]{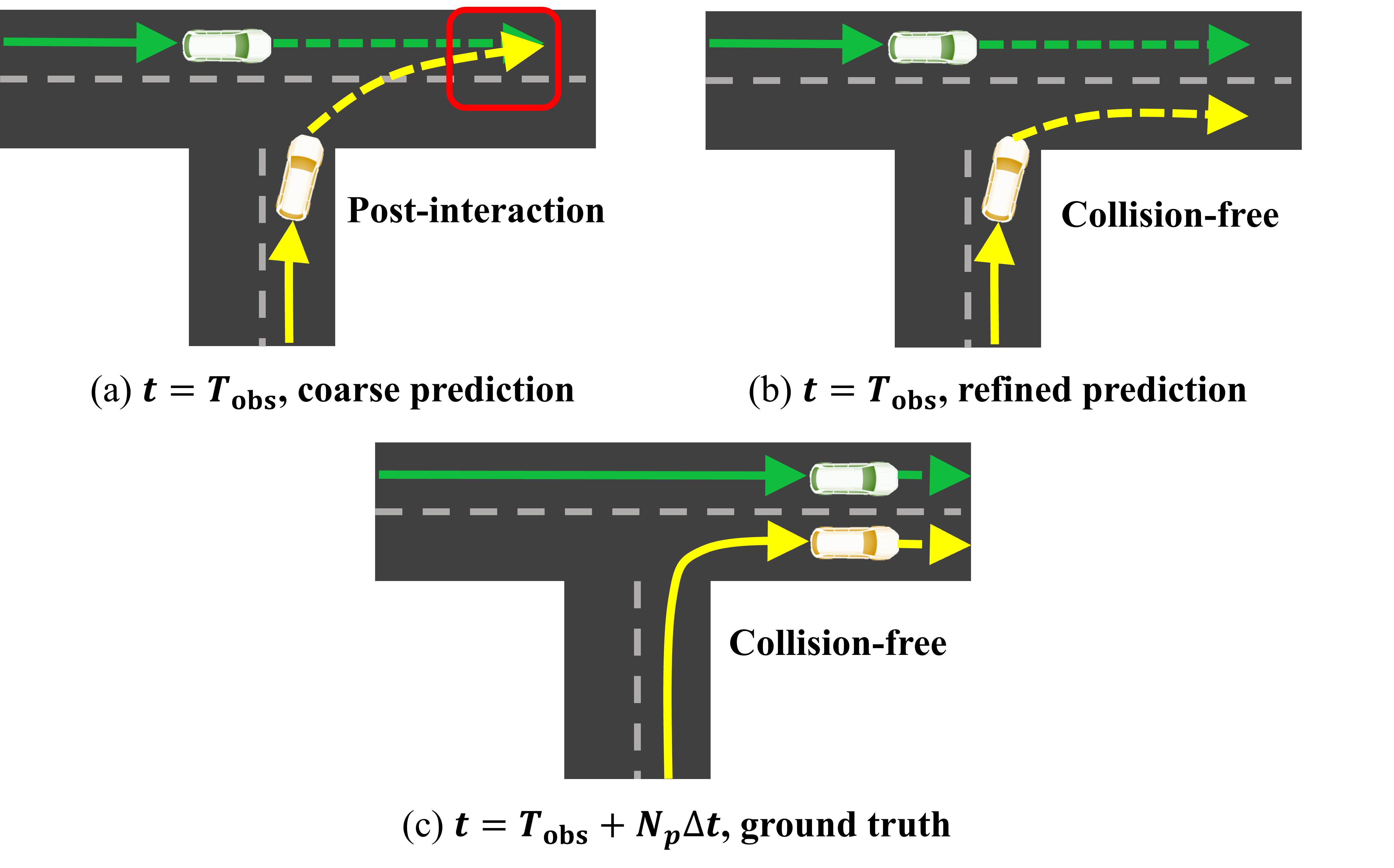}
    \caption{Illustration of trajectory predictions that consider post-interaction behavior. This study proposes a method, i.e., Pioformer,  for trajectory predictions of multiple agents. 
    (a) Current time step $t=T_\text{obs}$. We observe the post-interaction in the coarse predicted trajectory, which refers to the overlap between the predicted trajectories of two agents at future time step $t=T_\text{obs}+N_p\Delta t$. 
    (b) We use Pioformer to make predictions, which utilizes the post-interaction features embedded in coarse trajectories to refine the trajectory predictions.
    There are no collisions in the refined predicted trajectory.
    (c) Ground truth trajectories show that there are no collisions between the two agents at time step $t=T_\text{obs}+N_p\Delta t$. This indicates that our model can generate refined predictions that align well with the ground truth trajectory.  
    }
    \label{fig:post-interaction scenario}
\end{figure}

Interaction modeling has been crucial for trajectory predictions of multiple agents in strong-interactive traffic scenes \citep{ma2019trafficpredict, huang2023multimodal}. Social Force is a conventional method that uses attractive or repulsive forces to model the interactions among agents, while it cannot handle complex interactions \citep{helbing1995}. Recently, learning-based methods have been widely used for trajectory predictions, which use Recurrent Neural Networks (RNN) \citep{Alahi2016, Chandra2019}, Graph Convolutional Networks (GCNs) \citep{GCN, groupnet, HDGT} or Graph Attention Networks (GATs) \citep{GAT}, to learn the spatial and temporal interactions among multiple traffic agents \citep{mohamed2020, gao2020vectornet, Lanercnn, Ye2022, groupnet}. For instance, HiVT \citep{HiVT} extracts local context and models global interaction based on GAT and proposes translation-invariant scene representation and rotation-invariant spatial learning modules. Over the years, various variants of GNN have emerged \citep{groupnet, HDGT}, leading to continuous improvements in prediction performance. 

Modeling the interactions among agents for trajectory prediction of autonomous driving has been challenging due to the inherent uncertainty in agents' behavior. The interactions involved in the predicted trajectories of agents, also called post-interactions, have rarely been considered in trajectory prediction models.  The predicted trajectories sometimes include unexpected interactions such as traffic conflicts, lane deviations, or even collisions, which need more careful treatments.  It is unreasonable to directly remove those collided predictions since the crash could probably happen in some situations. Instead, we come to the idea of enhancing the trajectory prediction by utilizing those interactions involved in the predicted trajectories. Why can future interactions benefit trajectory prediction? The idea comes from the common sense that the participating agents are more likely to make proactive decisions to resolve the risks once unexpected interactions are predicted.  That is, those unexpected future interactions are less likely to turn into real risks and most trajectories would be collision-free and traffic rules-compliant at last. Moreover, we can infer that the future predictions can affect current decisions, and thus influence future motions of participating agents and then benefit the trajectory predictions. The interactions within the predicted trajectories, also called post-interactions, are expected to be helpful in trajectory predictions. Current methods for interaction modeling rarely consider the post-interaction and its potential impacts on trajectory predictions. In addition, standard GNN architectures for interaction modeling usually focus on low-order interactions and may fail to capture high-order interactions between the agents which are subtle and uncertain and directly affect the future motions of the agents. To address this, this study proposes a post-interactive multimodal trajectory prediction model, in which the high-order post-interaction features are learned to improve the prediction accuracy in strongly-interactive scenarios. Furthermore, we demonstrate how the multimodal trajectory can be integrated with the downstream planning under uncertainty and improve robustness for autonomous driving. 

\begin{figure*}[ht]
  \centering
  \includegraphics[width=1\textwidth]{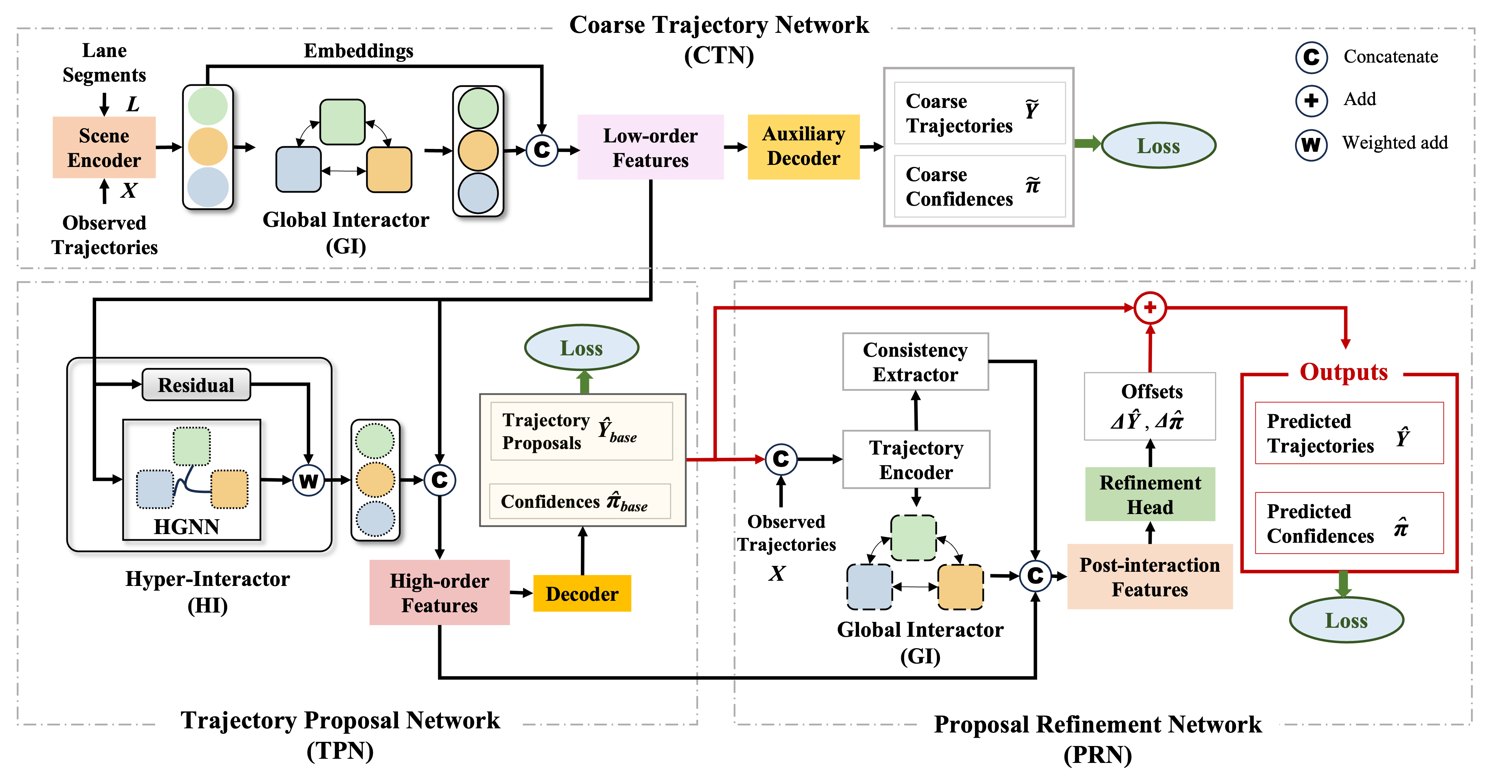}
  \caption{The architecture of Pioformer. The entire model can be divided into three networks: Coarse Trajectory Network (CTN), Trajectory Proposal Network (TPN) and Proposal Refinement Network (PRN). The top one is the CTN, which includes Scene Encoder, Global Interactor (GI), and Auxiliary Decoder, providing coarse predictions and low-order features (contextual information and simple pairwise interaction features). The bottom-left one is the TPN, which includes Hyper-Interactor (HI) and Decoder, generating trajectory proposals and corresponding confidences. HI leverages the low-order features to extract high-order post-interaction features (interactions beyond pairwise relationships) and refine the trajectories at the latent space. The bottom-right one is the PRN, which takes the trajectory proposals combined with observed trajectories as input. It further extracts post-interaction features of different trajectories and explores spatial-temporal consistency features for individual trajectories to generate offsets and refine initial proposals at the trajectory level. Both TPN and PRN are proposed post-interactive networks.}
  \label{fig:overall}
\end{figure*}

\subsection{Overview of this study}
\begin{figure*}[ht]
    \centering
    \includegraphics[width=\linewidth]{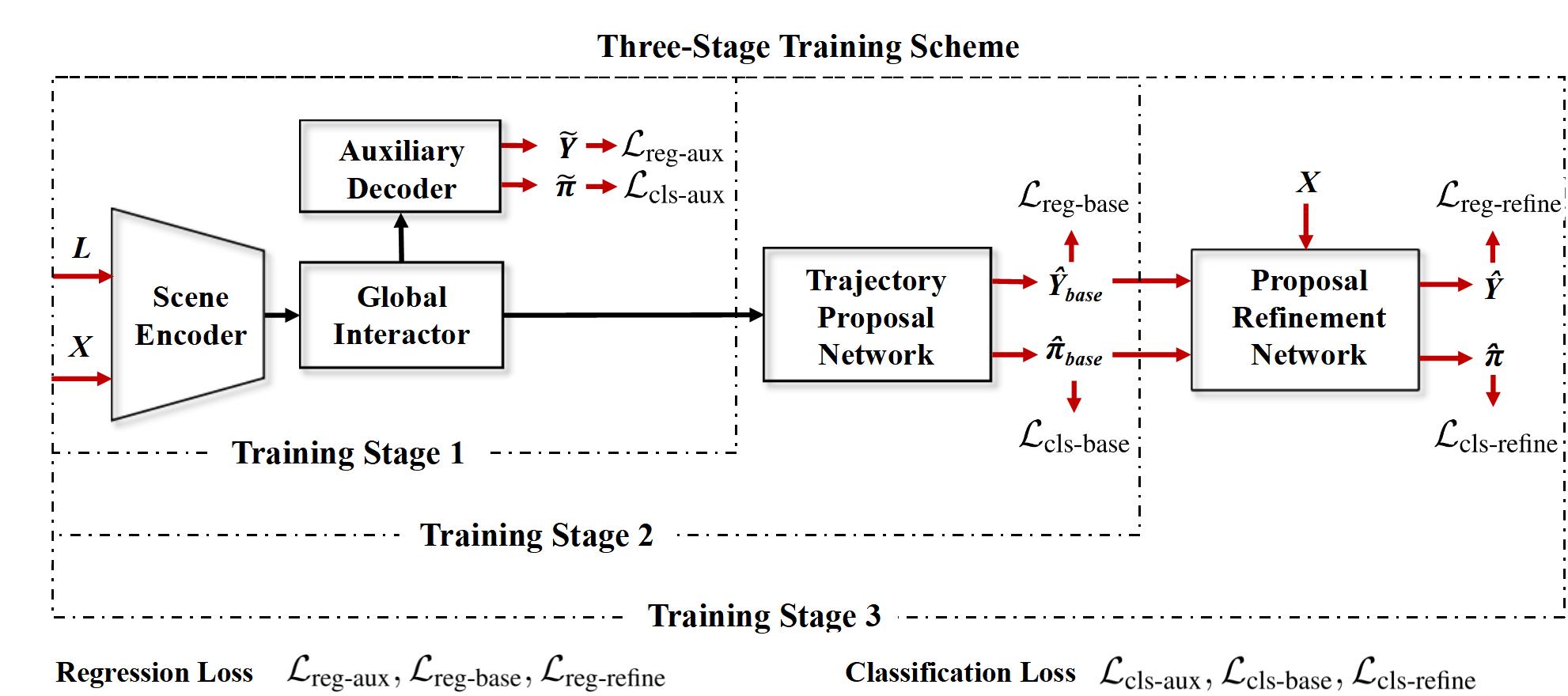}
    \caption{The overview of the three-stage training scheme. The three training stages are executed sequentially from left to right with each subsequent training stage reloading weights from the previous stage. In the first stage, we train CTN to achieve coarse trajectory prediction. In the second stage, we train CTN and TPN together, with the decoder in CTN acting as an auxiliary decoder while the decoder in TPN generates trajectory proposals. In the third stage, we train CTN, TPN and PRN together, with PRN refining the trajectory proposals to produce the final prediction. Each of the three networks has its corresponding trajectory regression loss and confidence classification loss.}
    \label{fig:three-stage}
\end{figure*}
\begin{figure} [t]
    \centering
    \includegraphics[width=0.65\linewidth]{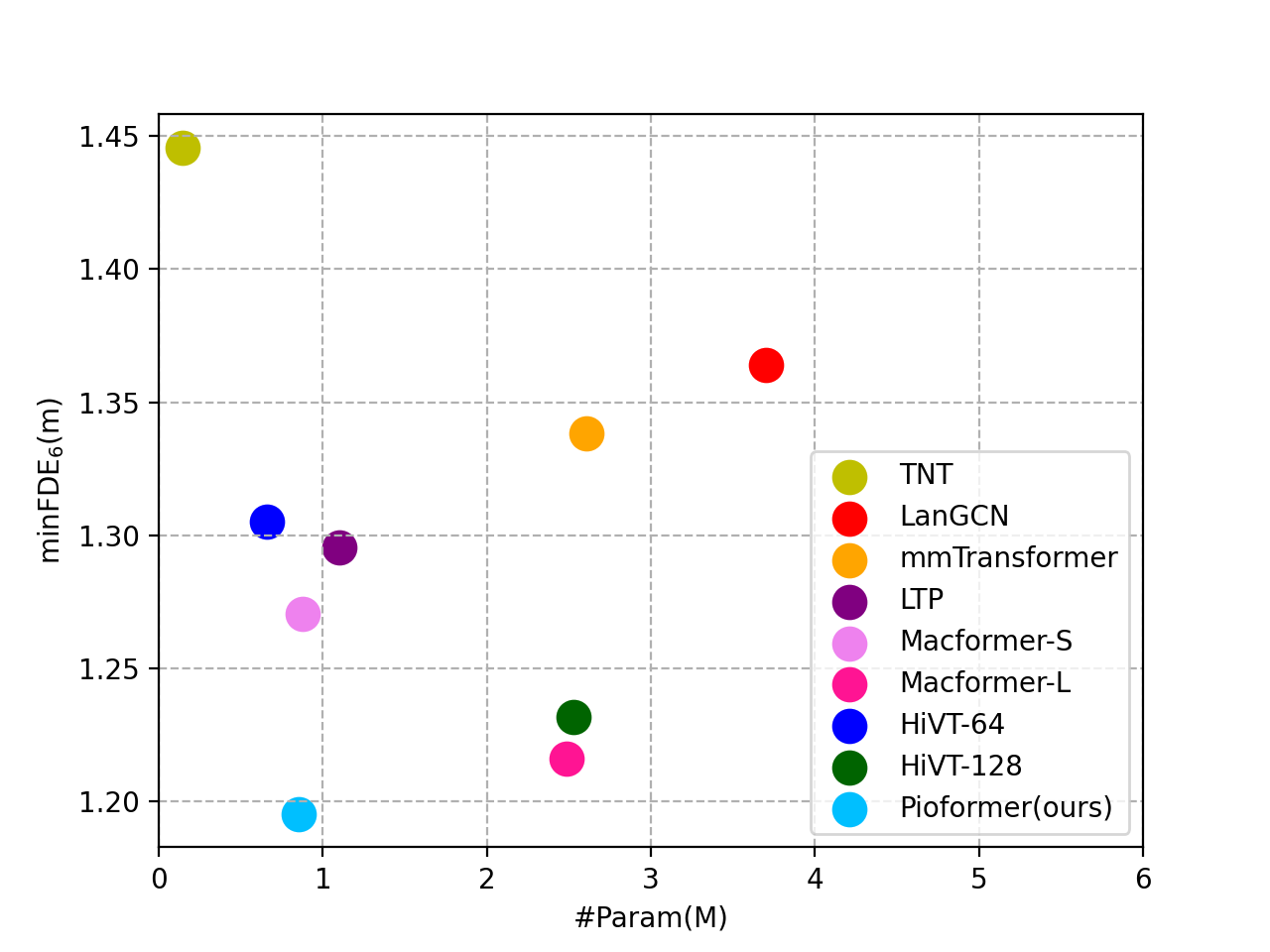}
    \caption{Overview of error-and-size trade-off for the task of trajectory prediction on \textit{Argoverse 1} leaderboard. With compact size, our models outperform most of the state-of-the-art models in prediction accuracy. Especially, the size of Pioformer is approximately one-third that of other models (e.g. HiVT-128 and Macformer-L) that achieve similar accuracy.
}
    \label{fig:trade-off}
\end{figure}
This study proposes a post-interactive multimodal trajectory prediction model for autonomous driving, \textit{i.e.}, Pioformer, which aims to model the post-interaction information among agents to enhance the trajectory predictions (see Fig. \ref{fig:post-interaction scenario}). As shown in Fig. \ref{fig:overall}, we employ a basic trajectory prediction model as a Coarse Trajectory Network, \textit{i.e.} backbone, to extract the local and global context and low-order interaction features. Next, a Hyper-Interactor module is built based on Hypergraph Neural Network (HGNN) to further learn the high-order post-interactions based on the previous low-order features. We construct the hypergraph topology based on the affinity of agents and then employ a mechanism similar to message-passing in ordinary GNNs to extract post-interaction features. The high-order features are then used to generate multimodal trajectory proposals and their confidences for each agent. The trajectory proposals are concatenated with the observed trajectories and used in Proposal Refinement Network to further extract the post-interaction features and correct unreasonable trajectories. Furthermore, a three-stage training scheme is designed to facilitate the training of Pioformer, as shown in Fig. \ref{fig:three-stage}. Extensive experimental results have shown the effectiveness of post-interaction features in enhancing the accuracy of multimodal trajectory prediction for multiple agents. Different from \citep{HiVT, laformer}, we propose a post-interaction aware trajectory prediction framework that employs hypergraphs to model high-order post-interaction patterns among vehicles and introduce a three-stage training scheme to facilitate the training process.
Our key contributions are summarized as follows:
\begin{itemize}
    \item We propose a multimodal trajectory prediction framework called Pioformer to leverage the post-interaction features to enable better predictions. A three-stage training scheme is introduced to facilitate the training.
    \item An HGNN-based Hyper-Interactor module is built to capture the high-order post-interaction features and adaptively refine the coarse trajectories in the latent space. We also design a Proposal Refinement Network, which can further utilize the post-interaction features and exploit spatial-temporal consistency to refine the initial trajectory proposals explicitly.
    \item Experiments demonstrate that our method can handle the trade-off between accuracy and model size (see Fig. \ref{fig:trade-off}) and outperform the baseline HiVT on the benchmark \textit{Argoverse 1} dataset, especially in strong-interactive scenarios.
\end{itemize}
\subsection{Organization of the Paper}
The subsequent sections of this paper are organized as follows. Section \ref{sec: related works} provides an overview of related works. Section \ref{sec: problem} presents the problem statement of trajectory prediction. Building on this foundation, Section \ref{sec: method} elaborates on the modules of Pioformer and introduces a three-stage training scheme. Section \ref{sec: experiments} presents extensive experimental results and analysis. The final section, Section \ref{sec: Conclusion} draws comprehensive conclusions of this paper.\\

\section{Related Works}\label{sec: related works}
This section introduces the related works on multimodal trajectory prediction, interaction modeling, coarse-to-fine trajectory prediction, and hypergraph neural networks.
\subsection{Multimodal Trajectory Prediction}
Due to the inherent uncertainties in driving behaviors and the surrounding environment, it is hard to generate one trajectory prediction that matches the ground truth well.  Multimodal trajectory prediction (MTP)  aims to generate multiple plausible and socially acceptable future predictions for each agent \citep{huang2023multimodal}, which has been widely studied in recent years \citep{gupta2018social, mohamed2020, gu2022stochastic}. 
Current MTP methods can be categorized into two groups,  i.e., the noise-based method and the anchor-based method. The noise-based methods typically use generative models to generate multimodal predictions, such as Social GAN \citep{socialgan} that is based on Generative Adversarial Network (GAN) \citep{GAN}, SocialVAE \citep{socialvae2022} that is based on Conditional Variational Autoencoder (CVAE) \citep{CVAE}, STGlow \citep{STglow} that is based on Normalising Flow (NF) \citep{NF}, and MID \citep{gu2022stochastic} that is based on Denoise Defusion Probabilistic Model (DDPM) \citep{DDPM}. However, those generative models often suffer from issues like training instability and mode collapse or may lack fine-grained details \citep{mode_collapse, kinematicsaware}. The anchor-based frameworks use anchors, such as the endpoints and the prototype trajectories, to guide the model to predict trajectories with controlled behaviors  \citep{huang2023multimodal,TNT}.  Compared with the noise-based methods, the anchor-based methods can generate more controllable predictions conditioned on prior knowledge and thus mitigate the mode collapse issues \citep{MultiPath, prophnet}. However,  if the anchors are not well defined, the unreachable endpoints or unreasonable prototype trajectories can degrade the performance of the anchor-based approaches \citep{PIH, DenseTNT}. Different from previous studies, our proposed model can be considered as a dynamic anchor-based method. Rather than relying on predefined anchors, we adopt the Trajectory Proposal Network to generate proposals that serve as anchors. Subsequently, these anchors are further refined by the Proposal Refinement Network, resulting in the final predicted trajectories.

\subsection{Interaction Modeling}
Modeling the interactions among agents has been essential to accurate trajectory prediction in complicated traffic scenes.  The Social Force Model is a popular approach used in crowd simulation that uses handcrafted features for interaction modeling, but it is hard to capture complex interaction behaviors. LSTM-based frameworks have been widely used for interaction modeling, such as Social LSTM \citep{Alahi2016}, TraPHic \citep{Chandra2019}, TrafficPredict \citep{ma2019trafficpredict}. However, the social pooling strategies in LSTM-based models cannot capture the important interactions between the agents that are located far from each other. Recently, GNNs have become increasingly popular in modeling the spatial-temporal dependencies among agents. For instance, VectorNet \citep{gao2020vectornet}  incorporates the vectorized scene context and agent dynamics for trajectory prediction based on graph neural networks. Social-STGCNN \citep{mohamed2020} models the interactions as a graph and proposes a weighted adjacency matrix in which the kernel function quantitatively measures pedestrian interactions. LaneGCN \citep{Lanegcn} extends graph convolutions with multiple adjacency matrices and along-lane dilation to effectively capture the lane graph's complex topology and long-range dependencies. LanceRCNN \citep{Lanercnn} learns a local lane graph representation per actor (LaneRoI) to encode its past trajectories and the local map topology. Recent works have expanded the conventional GCN and GAT to learn the interaction features. For instance, HDGT \citep{HDGT} utilizes a heterogeneous graph to model the traffic scenes and heterogeneous interactions. Most GNN-based methods extract low-order pairwise interactions between agents and often overlook the future interactions involved in agents' future trajectories, called post-interactions. 
Different from previous studies, we build a trainable HGNN to learn the high-order, group-wise interactions that efficiently address complex relational dynamics, especially post-interactions.
\textbf{
\subsection{Coarse-to-fine Trajectory Prediction}
}
Coarse-to-fine trajectory prediction refers to a hierarchical trajectory prediction approach where an initial coarse prediction is refined by incorporating more detailed features to improve accuracy.
The refinement module has been applied extensively in computer vision (CV). It is supported by its implementation in several tasks such as Cascade RCNN \citep{CascadeRCNN} for object detection and RefineNet \citep{refinenet} for semantic segmentation. These studies showcase the potential of the refinement module in improving prediction accuracy at a fine-grained level. In this task,  trajectory prediction refinement aims to correct the predicted trajectories. TPNet \citep{Fang2020TPNetTP} uses a two-stage framework for multimodal motion prediction of vehicles and pedestrians, in which the first stage generates a candidate set of future trajectories and the second stage performs classification and refinement on the proposals and selects the best one as the final prediction. MTR \citep{MTR} and MTR-A \citep{MTR-A} refine the trajectories by iteratively gathering fine-grained trajectory features. DESIRE \citep{DESIRE} refines the trajectories by incorporating the semantic context of scenes and the social interactions among agents. The Proposal Refinement Network of our proposed model is inspired by the  LAformer \citep{laformer}, and it aims to learn the trajectory offsets for correcting the trajectories. Different from this work, our approach extends upon it by incorporating post-interaction features, refining the confidence of each mode, and reusing an existing module to enhance model robustness. In addition to explicit refinement, we also refine the predicted trajectories in the high-dimensional latent space, aiming to move the initial trajectories to more plausible positions as much as possible.
\textbf{
\subsection{Hypergraph Neural Networks}
}
HGNN applications have been extensively adopted in diverse fields such as CV \citep{hgnn_cv}, natural language processing (NLP) \citep{hgnn_nlp}, multimodal learning \citep{hgnn_multi}, and data mining \citep{hgnn_dm}. HGNNs use hypergraphs to capture intricate, high-order dependencies between nodes that go beyond pairwise relationships \citep{HGNN, HGNN+}. Traffic scenarios involve numerous intricate high-order interactions among agents, such as interactions among multiple vehicles engaged in cooperative driving, as well as the sudden changing behaviors of vehicles in response to dynamic events. These can be effectively modeled using HGNN, where nodes represent agents and hyperedges are high-order interactions among agents \citep{groupnet, Dygroupnet}. Most research primarily focuses on modeling pairwise low-order interactions while neglecting high-order interactions, failing to comprehensively capture multi-faceted interactions among agents. We introduce HGNN to extract high-order post-interaction features and we make adaptive adjustments to accommodate the specific characteristics of autonomous driving and propose a corresponding three-stage training scheme to facilitate learning. Different from works such as GroupNet \citep{groupnet}, since our task focuses on vehicle trajectory prediction rather than pedestrian trajectory prediction. The interaction intensity between vehicles is significantly weaker than that between pedestrians. Therefore, we employ a single-scale hypergraph instead of a multi-scale hypergraph, reducing the computational complexity of the model.\\

\begin{figure} [t]
    \centering
    \includegraphics[width=0.4\linewidth]{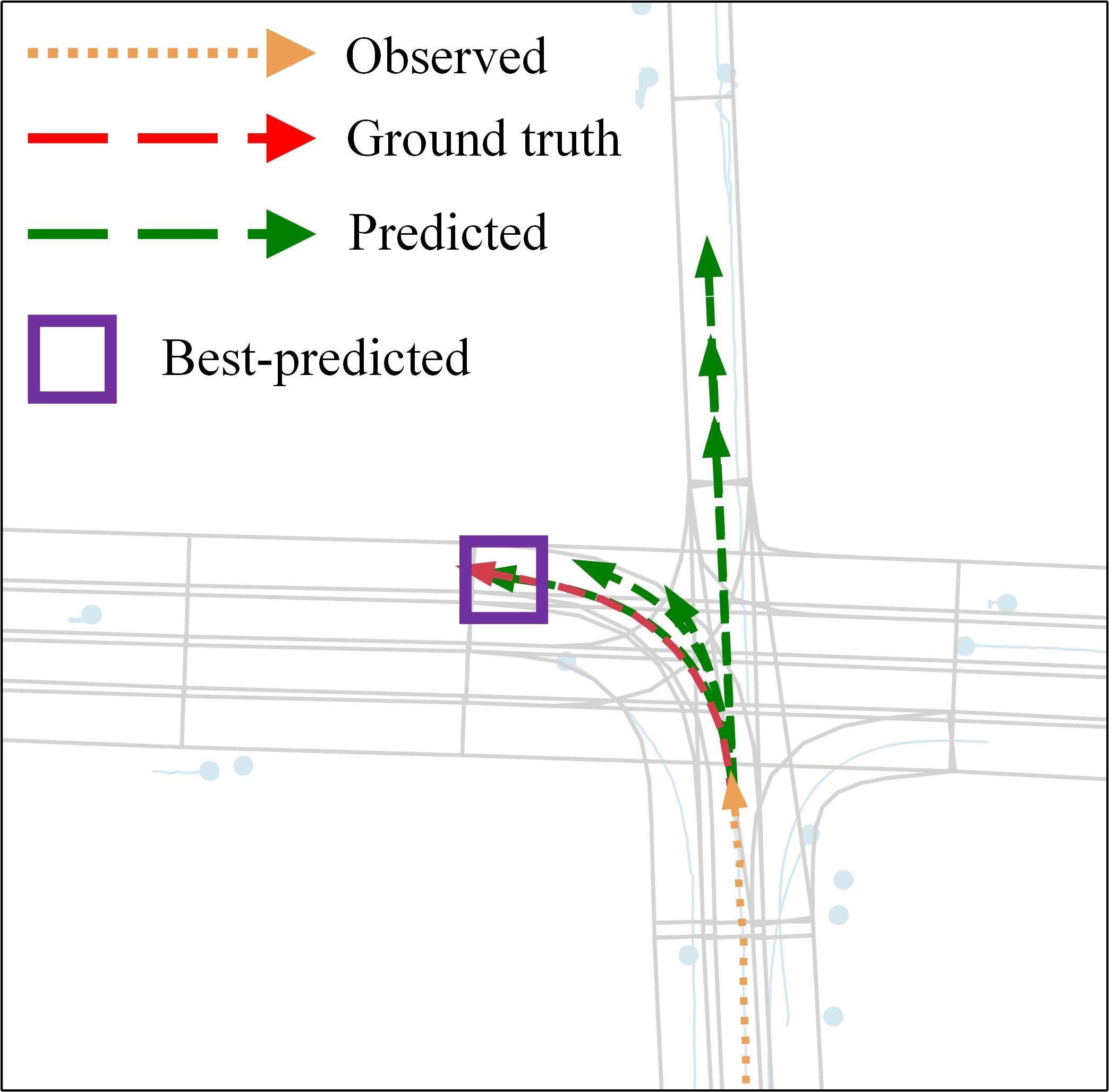}
    \caption{An example of multimodel trajectory prediction in a traffic scenario. We predict multiple possible trajectories based on the observed trajectories of agents and lane information.  
}
\label{fig:problem}
\end{figure}
\section{Problem Formulation}\label{sec: problem}
In this section, we formulate the multimodal trajectory prediction problem as a sequence-to-sequence model, and our goal is to generate multiple predictions for each agent based on the observed trajectories. 
\subsection{Multimodal Trajectory Prediction}
We focus on a multimodal trajectory prediction model, which aims to generate a distribution of future trajectories for each agent. The trajectory prediction model leverages the observed trajectories of multiple agents and lane segment information to predict the trajectory distributions, as shown in Fig. \ref{fig:overall}.  
Then, the best-predicted trajectory is selected from the trajectory distribution by minimizing the prediction errors, as shown in Fig. \ref{fig:problem}. That is, the trajectory mode that is located closest to the ground truth trajectory is chosen as the best-predicted one.    

We denote the observed state of the agent $i$ at time step $t$ as $\mathbf{x}_i^t$, the observed trajectory is written as $\mathbf{X}=\left\{\mathbf{x}_i^t\right\}$, where $i \in \left\{1,2,\cdots, M\right\}$ and $t \in \left\{1,2,\cdots, T_\text{obs}\right\}$, $M$ is the number of agents, $T_\text{obs}$ is the observed time steps. Similarly, we denote the predicted state of the agent $i$ at time step $t$ of the mode $k$ as $\mathbf{\hat{y}}_{i,k}^t$, and the predicted trajectories $\mathbf{\hat{Y}}=\left\{\mathbf{\hat{y}}_{i,k}^t\right\}$, where $t \in \left\{T_\text{obs}+1,\cdots, T_{\text{pred}}\right\}$, $k \in \left\{1,2, \cdots, K\right\}$, and $K$ is the number of prediction modes. Then, the trajectory prediction problem can be represented as,
\begin{equation}
\textbf{f}: (\mathbf{X},L) \rightarrow \mathbf{\hat{Y}}
\end{equation}
where $\textbf{f}$ is the multimodal trajectory prediction model, $\mathbf{X}$ denotes the observed trajectories, $L$ is the lane segment encodings, $\mathbf{\hat{Y}}$ denotes the predicted trajectories. 

To find the best-predicted trajectory from $K$ modes, the accumulated prediction error of each prediction mode is calculated based on the ground truth trajectory. The trajectory mode that has the minimum accumulated prediction error is selected as the best-predicted trajectory,
\begin{equation}
 k_i^*=  \underset{k}{\operatorname{argmin}}\sum_{t=T_{\text{obs}}+1}^{T_{\text{pred}}}\left\|\mathbf{y}_i^t-\hat{\mathbf{y}}_{i, k}^t\right\| \quad k \in\{1,2, \ldots, K\},
\label{eq:p1.1}
\end{equation}
where $k_i^*$ is the best-predicted mode of agent $i$,  $\mathbf{y}_i^t$ is the ground truth state of agent $i$ at time step $t$, $\mathbf{\hat{y}}_{i,k}^t$ is the predicted state of agent $i$ of mode $k$ at time step $t$. 

\textbf{\subsection{Scene Representation}}
A translation-invariant scene representation is employed in this study, similar to \citep{HiVT}. We represent the agent trajectories and lane segments using relative positions to ensure translation invariance. Therefore, the entire scene can
be represented as a set of vectors. 
\begin{itemize}
    \item \textbf{Agent}: the position of agent \(i\) at time \(t\) can be represented as \(\mathbf{p}_i^t=\left(p_{i, x}^t, p_{i, y}^t\right)\). The trajectory of agent \(i\) can then be represented as \(\left\{\mathbf{p}_i^t -\mathbf{p}_i^{t-1}\right\}_{t=1}^{T_\text{obs}}\). 
    \item \textbf{Lane}: a lane segment with the starting point $\mathbf{p}_l^s$ and the ending point $\mathbf{p}_l^e$ can be represented as $\mathbf{p}_l^e-\mathbf{p}_l^s$. 
    \item \textbf{Scene elements}: the spatial relationship between scene elements can also be represented as relative position vectors, and $\mathbf{p}_j^t -\mathbf{p}_i^t$ denotes the position vector of agent $j$ relative to agent $i$ at time step $t$, and $\mathbf{p}_l^s-\mathbf{p}_i^t$ denotes the position vector of lane segment $l$ related to agent $i$ at time step $t$. 
    \item \textbf{Local region}: like \citep{HiVT}, we divide the scene into local regions, selecting one agent as the central agent, which is located in the center of the local region. We use the latest trajectory segment $\mathbf{p}_i^{T_\text{obs}} -\mathbf{p}_i^{T_\text{obs}-1}$ of central agent $i$ as the reference vector and all vectors are rotated according to the orientation angle $\theta_i$ of the reference vector. The rotation matrix is represented as $\mathbf{R}^{\top}_i$ matrix parameterized by $\theta_i$. 
\end{itemize}

\section{Methodology}\label{sec: method}
This section introduces our model, Pioformer, and three main parts, including Coarse Trajectory Network, Trajectory Proposal Network, and Proposal Refinement Network. 
\textbf{\subsection{Overall Framework}} 
Fig. \ref{fig:overall} illustrates our model, which can be divided into three networks, i.e., Coarse Trajectory Network (CTN), Trajectory Proposal Network (TPN) and Proposal Refinement Network (PRN). The top one is the CTN, which includes Scene Encoder, Global Interactor (GI), and Auxiliary Decoder, providing coarse predictions and low-order interaction features (contextual information and simple pairwise interaction features). The bottom-left one is the TPN, which includes Hyper-Interactor (HI) and Decoder, generating trajectory proposals and corresponding trajectory confidences based on low-order interaction features. In particular, the Hyper-Interactor module in TPN is built to extract high-order post-interaction features (interactions beyond pairwise relationships) leveraging the low-order interaction features. The bottom-right one is the PRN, which takes the trajectory proposals from TPN and observed trajectories as input and output refined trajectories. We note that PRN is designed to further extract post-interaction features of multiple agents and explore spatial-temporal consistency features to refine trajectory proposals.
\textbf{\subsection{Coarse Trajectory Network (CTN)}}
CTN aims to encode the observed trajectories and context information and then decode coarse-grained trajectory predictions while providing low-order interaction features for subsequent networks. In this paper, we adopt HiVT as our CTN, which consists of three components: Scene Encoder, Global Interactor and Auxiliary Decoder. This section will provide a brief introduction to these modules, for detailed information please refer to \citep{HiVT}. \\
\subsubsection{\textbf{Agent-Agent Interaction}} In each local region, we extract the features of the central agent \(i\), denoted as $\mathbf{z}_i^t$, as well as the features of any neighboring agents \(j\) w.r.t. the agent $i$, denoted as $\mathbf{z}_{i j}^t$, at time step \(t\) :
\begin{equation}
    \mathbf{z}_i^t=\phi_{\text {center }}\left(\left[\mathbf{R}_i^{\top}\left(\mathbf{p}_i^t-\mathbf{p}_i^{t-1}\right), \mathbf{a}_i\right]\right),\label{eq:1}
\end{equation}
\begin{equation}
    \mathbf{z}_{i j}^t=\phi_{\text{nbr}}\left(\left[\mathbf{R}_i^{\top}\left(\mathbf{p}_j^t-\mathbf{p}_j^{t-1}\right), \mathbf{R}_i^{\top}\left(\mathbf{p}_j^t-\mathbf{p}_i^t\right), \mathbf{a}_j\right]\right),\label{eq:2}
\end{equation}
where \(\phi_{\text {center }}(\cdot)\) and \(\phi_{\text{nbr}}(\cdot)\) are both MLP blocks,  \(\mathbf{a}_i\) denotes the semantic attributes of agent \(i\). We then utilize a rotation-invariant cross-attention mechanism to obtain spatial interaction features among the agents. Subsequently, we need to capture temporal dependencies by aggregating each agent’s embeddings in chronological order using a Transformer module to obtain spatio-temporal interaction embeddings.
\subsubsection{\textbf{Agent-Lane Interaction}} We need to leverage features from the surrounding lane segments to guide the future motion of the agents:
\begin{equation}
    \mathbf{z}_{il}=\phi_{\text {lane }}\left(\left[\mathbf{R}_i^{\top}\left(\mathbf{p}_l^e-\mathbf{p}_l^s\right), \mathbf{R}_i^{\top}\left(\mathbf{p}_l^s-\mathbf{p}_i^{T_{\text{obs}}}\right), \mathbf{a}_l\right]\right),\label{eq:5}
\end{equation}
where $\mathbf{z}_{il}$ represents the embedding between the agent \(i\) and the lane segment \(l\), \(\phi_{\text {lane }}(\cdot)\) is an MLP block,  \(\mathbf{a}_l\)  is the semantic attributes of the lane segment. Similarly, the agent-lane attention is calculated using the cross-attention module. The final output $\boldsymbol{\psi}^{\text{local}}_i$ , referred to as the local embedding of the agent $i$ is the fusion of spatial-temporal interaction features in the local region.

\subsubsection{\textbf{Global Interactor} (GI)} This module aims to further extract long-range dependencies and establish geometric relationships between local regions to bridge the differences between their respective coordinate systems. For agents $i$ and $j$, we continue to employ an MLP block $\phi_{\text {rel}}(\cdot)$ to extract embeddings representing the pairwise interactions between the two agents, denoted as $\mathbf{s}_{i j}$:
\begin{equation}
\mathbf{s}_{ij}=\phi_{\text {rel }}\left(\left[\mathbf{R}_i^{\top}\left(\mathbf{p}_j^{T_{\text{obs}}}-\mathbf{p}_i^{T_{\text{obs}}}\right), \cos \left(\Delta \theta_{ij}\right), \sin \left(\Delta \theta_{ij}\right)\right]\right),
\end{equation}
where $\mathbf{p}_j^{T_{\text{obs}}}-\mathbf{p}_i^{T_{\text{obs}}}$ represents the differences between agent $i$’s and agent $j$’s coordinate frames, and $\Delta \theta_{i j} $ denotes $\theta_{j} - \theta_{i}$. 
Subsequently, the cross-attention mechanism is employed again to further obtain the global embedding $\boldsymbol{\psi}^{global}_i$.
Up to now, we have successfully captured both local and long-range dependencies from the agent's observed trajectory and obtained low-order interaction features. 

\subsubsection{\textbf{Auxiliary Decoder}} We employ the Laplace Mixture Model (LMM)  to parameterize the distribution of future trajectories, and the predicted multimodal trajectory distribution of agent $i$ is denoted as $\mathbf{\tilde{Y}}_i$:
\begin{equation}
\mathbf{\tilde{Y}}_i = \sum_{k=1}^K \tilde{\pi}_{i, k}\mathfrak{L}\left(\boldsymbol{\tilde{\mu}}_{i, k}, \mathbf{\tilde{b}}_{i, k}\right), \label{eq:9}
\end{equation}
where $\mathfrak{L}(\cdot,\cdot)$ denotes Laplace distribution, the predicted probability that agent \(i\) adopts mode \(k\)  is \(\tilde{\pi}_{i, k}\), the predicted location and uncertainty parameters of each Laplace component are denoted as  \(\boldsymbol{\tilde{\mu}}_{i, k}\) and \(\mathbf{\tilde{b}}_{i, k}\). The decoder takes the previous embeddings as input and employs three separate MLP blocks to generate the three parameters of the Laplace distribution. We note that this auxiliary decoder is only employed during the training process and is not used in the inference stage. 

\subsection{Trajectory Proposal Network (TPN)}

\begin{figure*}[ht]
    \centering
    \includegraphics[width=0.9\linewidth]{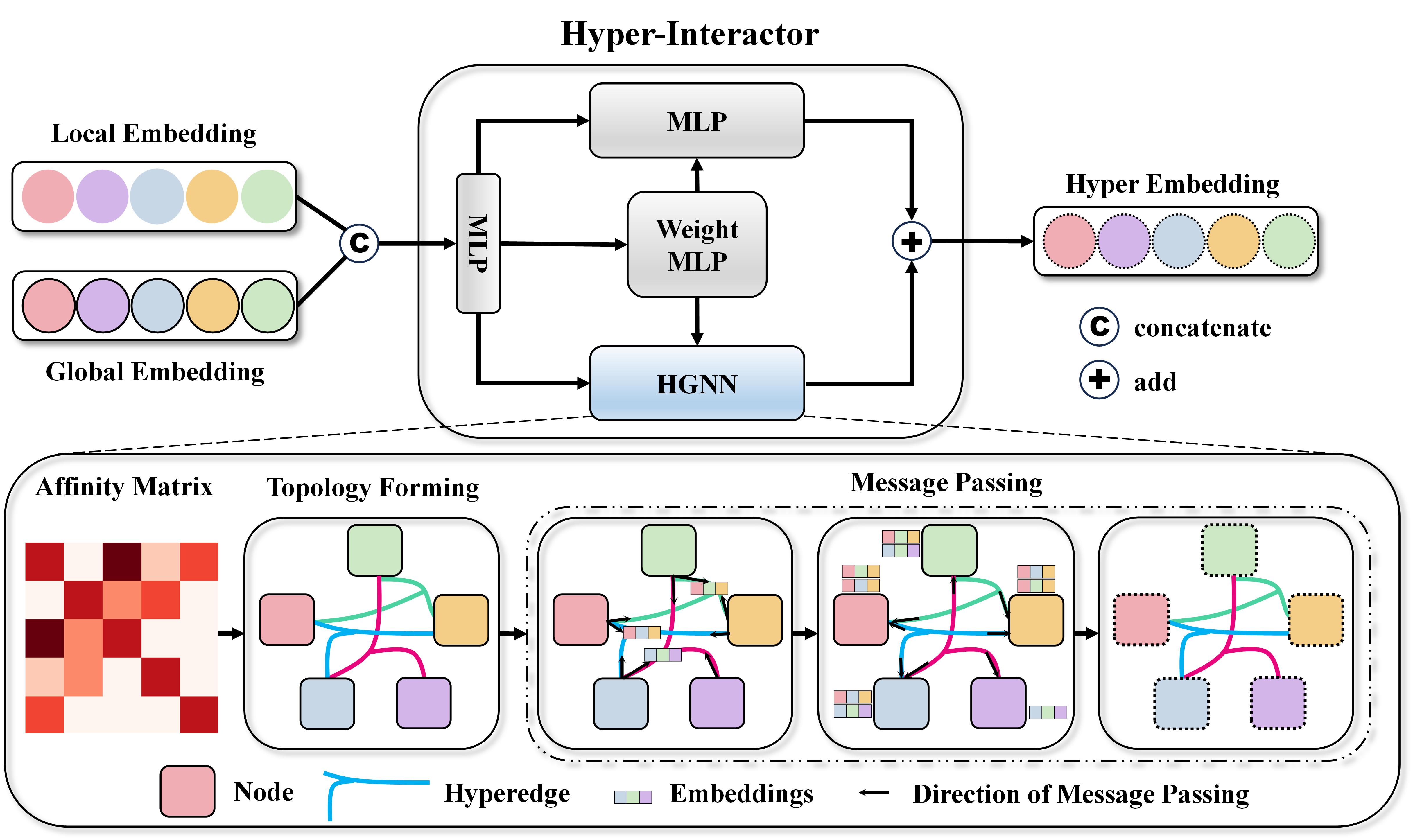}
    \caption{The architecture of the Hyper-Interactor module. The core of Hyper-Interactor is an HGNN, which is aimed at capturing high-order interaction features. The hypergraph topology is adaptively constructed by calculating the affinity between agent nodes, which is then used for message propagation. Additionally, an MLP block is utilized as a residual branch to provide low-order interaction features. The high-order and low-order features are then integrated in a weighted manner to obtain the final output. }
    \label{fig:hyper-interactor}
    \vspace{-3mm}
\end{figure*}

TPN includes Hyper-Interactor (HI) and Main Decoder, which aims to model high-order interactions using HGNNs based on the low-order interaction features provided by the CTN and output multimodal trajectory proposals and related trajectory confidences, (see Fig. \ref{fig:hyper-interactor}). HI takes previous low-order embeddings as input, which are subsequently decoded into real trajectories in CTN. This implies that these embeddings represent the future trajectories in a high-dimensional space. We then use an MLP block to fuse the previous embeddings, thereby generating features relevant to agents' future trajectories. With these embeddings, we can proceed with post-interaction modeling and learning. 

\begin{figure} [t]
    \centering
    \includegraphics[width=0.55\linewidth]{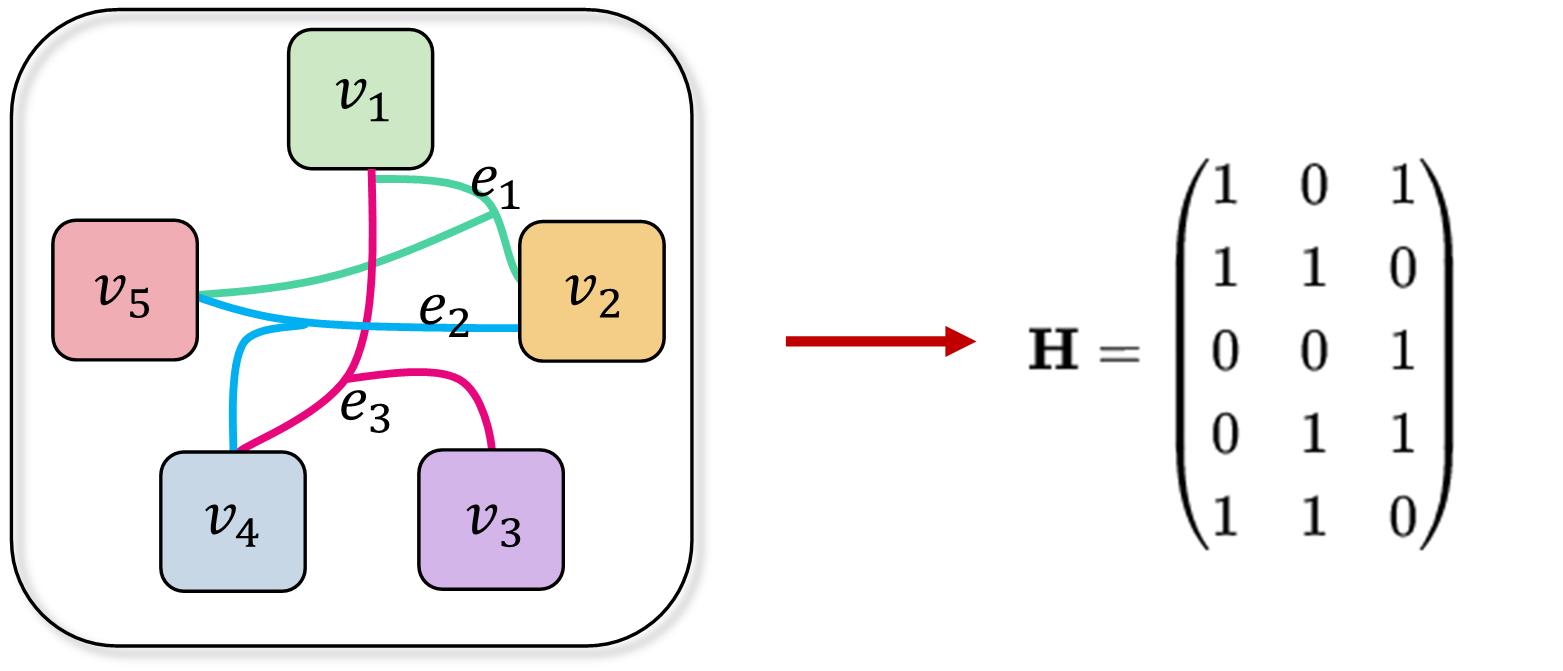}
    \caption{An example of a hypergraph. It consists of 5 vertices represented by rounded rectangles and 3 hyperedges represented by curves. The rows and columns of the incidence matrix $\mathbf{H}$ represent vertices and hyperedges,  with the elements indicating whether the vertex belongs to the hyperedge. For instance, hyperedge $e_1$ connects $v_1$, $v_2$ and $v_5$, then the first column of the matrix would be represented as $(1, 1, 0, 0, 1)$.  
}
\label{fig:hypergraph}
\end{figure}

\subsubsection{\textbf{Hyper-Interactor (HI)}}

We use hypergraph neural networks instead of the graph neural networks to learn the high-order post-interaction features, facilitating an in-depth exploration of both low-order and high-order interaction features.

\textbf{Hypergraph}. Mathematically, a hypergraph is a mathematical structure that extends the concept of a simple graph. In a simple graph, the edges represent pairwise connections between nodes. However, in a hypergraph, the hyperedges are capable of connecting any number of nodes, thus allowing for a more flexible and expressive representation of relationships among nodes. Formally, a hypergraph can be represented as:  $\mathcal{G}=(\mathcal{V}, \mathcal{E})$, where $\mathcal{V} = \{v_1, v_2, ... v_{N_V}\}$ denotes the vertex set which contains $N_V$ vertices, $\mathcal{E} = \{e_1, e_2, ... e_{N_E}\}$ denotes the hyperedge set which contains $N_E$ hyperedges. The topological structure of a hypergraph can be represented using an incidence matrix $\mathbf{H} \in \mathbb{R}^{|\mathcal{V}| \times|\mathcal{E}|}$ which is a binary matrix with entries defined as
\begin{equation}
h(v, e)= \begin{cases}1, & \text { if } v \in e \\ 0, & \text { if } v \notin e,\end{cases}
\end{equation}
where $v \in \mathcal{V}$ and $e \in \mathcal{E}$ represent vertex and hyperedge respectively \citep{HGNN}. 
To make it clear,  we provide an example in Fig. \ref{fig:hypergraph}. It consists of 5 vertices represented by rounded rectangles and 3 hyperedges represented by curves. The rows and columns of the incidence matrix $\mathbf{H}$ represent vertices and hyperedges,  with the elements indicating whether the vertex belongs to the hyperedge. For instance, hyperedge $e_1$ connects $v_1$, $v_2$ and $v_5$, then the first column of the matrix would be represented as $(1, 1, 0, 0, 1)$.

\textbf{Hypergraph Neural Netwowrks}. HGNNs are a class of neural networks that leverage the capacity of hypergraphs to capture complex, high-order dependencies among nodes, going beyond pairwise relationships. HGNNs employ an iterative message-passing strategy for information propagation, similar to GNNs, facilitating the integration of the local and global hypergraph structures in the update of node features. During each message-passing iteration, nodes engage in information exchange with their connected hyperedges. In turn, the hyperedges transmit this rich information to other connected nodes. 

The initial step for constructing an HGNN is to define the topology of the hypergraph. As interactions between agents in traffic scenarios are unknown, it is hard to define the topology. Therefore, following GroupNet \citep{groupnet}, we can use an affinity matrix to calculate the correlation between different agents to identify a suitable topology for building a hypergraph. With cosine similarity adopted as a metric, the entries of the affinity matrix  $\mathbf{A} \in \mathbb{R}^{M \times M}$ are defined as
\begin{equation}
\mathbf{A}_{i, j} = \frac{{\boldsymbol{\tau}^{\top}_i \boldsymbol{\tau}_j}}{{\left|\left|\boldsymbol{\tau}_i \right|\right|\left|\left|\boldsymbol{\tau}_j \right|\right|}},\label{eq:8}
\end{equation}
where \(\boldsymbol{\tau}_i\) is the future trajectory feature of agent $i$. We can further use the affinity matrix to search for high-density submatrices. We assume that the agent nodes located on the same hyperedge should have a higher affinity with each other. The search for hyperedges with a predetermined number of connected nodes $S$ can be viewed as solving an optimization problem. We aim to find an optimal hyperedge that can maximize the norm of the affinity matrix, 
\begin{equation}
\begin{aligned}
& e_i=\arg \max _{\Omega \subseteq \mathcal{V}}\left\|\mathbf{A}_{\Omega, \Omega}\right\|_{1,1}, \\
& \text { s.t. }|\Omega|=S, v_i \in \Omega, i=1, \ldots, M,
\end{aligned}\label{eq:8}
\end{equation}
where $e_i$ is the optimal hyperedge of agent $i$, $\left\|\cdot\right\|_{1,1}$ is an entrywise matrix norm that measures the sum of the absolute values of all elements in the matrix, and $|\Omega|$ is the cardinal of the vertex set $\Omega$. The first condition in the formula imposes a limitation on the number of agents on each hyperedge. The following condition guarantees that no agent is left out. We can use an exhaustive approach to solve the optimization problem directly when the number of agents in practical traffic scenarios is small, usually in the range of a few dozen.

With the topological structure of the hypergraph, we can utilize it for message-passing. The message-passing process in a hypergraph is similar to that in a simple graph. 
During the node-to-hyperedge phase, node embeddings are aggregated into hyperedges, which can be represented as:
\begin{equation}
\mathbf{e}_i= \phi_{\text{hyperedge}}\left(\sum_{v_j \in e_i} \mathbf{v}_j\right).\label{eq:8}
\end{equation}
where the embedding of each node is denoted as $\mathbf{v}_j$,
$\mathbf{e}_i$ is the embedding of hyperedge $e_i$ and $\phi_{\text{hyperedge}}\left(\cdot\right)$ is an MLP to obtain the interaction embeddings. 

During the hyperedge-to-node phase, nodes update their embeddings by incorporating the interaction embeddings learned from the hyperedges. The updated embedding, $\mathbf{v}_i$,  can be obtained through:
\begin{equation}
\mathbf{v}_i \leftarrow \phi_{\text{node}}\left(\left[\mathbf{v}_i, \sum_{e_j \in \Omega_i} \mathbf{e}_j\right]\right),
\end{equation}
where $\Omega_i$ is a set of hyperedges containing node $i$, and $\phi_{\text{node}}\left(\cdot\right)$ is an MLP to obtain updated embeddings. This message-passing process facilitates the iterative refinement of node embeddings.  

Consequently, the HGNN yields embeddings denoted as \(\mathbf{r}^{\text{high}}\) that represent the high-order interactions of various agents and their latent impact in the future.

In many traffic scenarios, especially in straight-line situations with few surrounding agents, the post-interactions among agents are relatively slight. In such scenarios, the initially predicted trajectories may already meet the requirements. Employing the HGNN for processing may generate overfitting issues and result in undesired lane changes. Thus, we use a simple MLP  $\phi_{\text{s}}(\cdot) $ as a residual block to leverage the embedding $\boldsymbol{\tau_i}$ to compute low-order interaction features:
\begin{equation}
\mathbf{r}_i^{\text{low}}=\phi_{\text {s }}\left(\boldsymbol{\tau}_i\right).
\end{equation}
Where $\mathbf{r}_i^{\text{low}}$ is 
the low-order interaction features. 

To adaptively incorporate the high-order and low-order interaction features,  a weighting mechanism is adopted to incorporate the results of the HGNN and the simple MLP block:
\begin{equation}
\boldsymbol{\omega}_i=\sigma\left(\phi_{\text {weight}}\left(\boldsymbol{\tau}_i\right)\right),\label{eq:11}
\end{equation}
\begin{equation}
\boldsymbol{\psi}_i^{\text{hyper}}=\mathbf{r}_i^{\text{high}} \odot \boldsymbol{\omega_i}+ \mathbf{r}_i^{\text{low}} \odot \left(1-\boldsymbol{\omega}_i\right),\label{eq:13}
\end{equation}
where \(\phi_{\text {weight }}(\cdot)\) is an MLP block, \(\sigma\left(\cdot\right)\) denotes the sigmoid function,  \(\odot\) denotes the element-wise product. $\boldsymbol{\psi}_i^{\text{hyper}}$ is the weighted post-interaction embedding of agent $i$. This weighting mechanism enables the flexibility to selectively leverage the output of the MLP and the output of the HGNN to accommodate the interactions in different traffic scenarios.\\

\subsubsection{\textbf{Main Decoder}}
Hyper-Interactor module captures the dynamic nature of high-order post-interactions, enabling refinement of the original trajectory embeddings based on the interactions between the initially predicted trajectories in the latent space. We subsequently employ a decoder to generate trajectory proposals, which serves as the main decoder. Similar to the auxiliary decoder, the trajectory proposal $\mathbf{\hat{Y}}_{\text{base},i}$ generated by the main decoder can be expressed as:
\begin{equation}
\mathbf{\hat{Y}}_{\text{base},i} = \sum_{k=1}^K \hat{\pi}_{\text{base}, i, k}\mathfrak{L}\left(\boldsymbol{\hat{\mu}}_{\text{base},i, k}, \mathbf{\hat{b}}_{\text{base},i, k}\right), \label{eq:9}
\end{equation}
where \(\hat{\pi}_{\text{base}, i, k}\), \(\boldsymbol{\hat{\mu}}_{\text{base},i, k}\) and \(\mathbf{\hat{b}}_{\text{base}, i, k}\) are the predicted parameters associated to the trajectory proposals. The main decoder leverages the high-order features along with the previously obtained low-order features to generate more accurate trajectory proposals, which serve as anchor trajectories for the Proposal Refinement Network.
\subsection{Proposal Refinement Network (PRN)}
\label{sec:PR}
PRN takes the trajectory proposals and observed trajectories as input, and aims to learn the post-interaction features among multiple agents and explores spatial-temporal consistency features to generate offsets and refine trajectory proposals. 
\subsubsection{\textbf{Post-interaction Embedding}}
The observed trajectory $\mathbf{X}_i$ is concatenated with the trajectory proposal $\hat{\mathbf{Y}}_{\text{base},i}$, and then sent to a trajectory encoder to represent the trajectory embedding $\mathbf{\eta}_i$:
\begin{equation}
\mathbf{\eta}_i=\mathbf{f}_{\text{encode}}\left(\left[\mathbf{X}_i, \mathbf{\hat{Y}}_{\text{base}, i}\right]\right),\label{eq:14}
\end{equation}
where $\mathbf{f}_{\text{encode}}\left(\cdot\right)$  denotes the trajectory encoder that is built with a bi-directional GRU \citep{GRU}. 

The trajectories from the previous decoder may have misalignment or inconsistency with observed trajectories, such as sudden acceleration, excessively large steering angles, etc. 
To address this, we employ a three-layer MLP block to build the consistency extractor $\phi_{\text {consist }}\left(\cdot\right)$ based on the trajectory embedding \(\mathbf{\eta}_i\) ,
\begin{equation}
 \boldsymbol{\psi}^\text {consist}_i=\phi_{\text {consist }}\left(\mathbf{\eta}_i\right),    
\end{equation}
where $\boldsymbol{\psi}^\text {consist}_i$ denotes the trajectory consistency embedding, which is expected to capture the spatial-temporal consistency features of the entire trajectory. This can be considered as interactions between future trajectories and observed trajectories, which enables a better understanding of the agent's interaction behavior. The trajectory consistency embedding is used in the subsequent trajectory refinement process in order to improve reasonableness and adherence to physical constraints. 

To further explore post-interaction features, we continue to leverage trajectory embeddings $\mathbf{\eta}_i$ over the entire time horizon. Due to the use of relative coordinates to represent contextual scenes, additional information about the geometric relationships between agents must be provided following a straightforward encoding process. 
Thus, the trajectory embedding $\mathbf{\eta}_i$ is also fed into Global Interactor module to obtain the spatial-temporal post-interaction features: 
\begin{equation}
 \boldsymbol{\psi}^{\text {post-int}}_i=\mathbf{f}_{\text{global}}\left(\mathbf{\eta}_i\right),    
\end{equation}
where $\boldsymbol{\psi}^{\text {post-int}}_i$ represents post-interaction embeddings, $\mathbf{f}_{\text{global}}\left(\cdot\right)$ is the Global Interactor. 
By reusing the previous modules, we can reduce the number of parameters and allow Global Interactor to be trained sufficiently.
\subsubsection{\textbf{Trajectory Output}}
We have obtained multiple embeddings, including $\boldsymbol{\psi}_i^\text {local},\boldsymbol{\psi}_i^\text {global},\boldsymbol{\psi}_i^{\text {hyper}},\boldsymbol{\psi}_i^\text {consist},\boldsymbol{\psi}_i^{\text {post-int}}$, which are fed into the Refinement Head in PRN to generate the trajectory offsets for correcting the trajectory proposals. Considering the uncertainties involved in multiple trajectory modes, the PRN uniformly corrects the trajectory offsets and confidences of each mode. Here,  
a three-layer MLP block \(\phi_\text {refine-traj}\left(\cdot\right)\) is adopted to generate position offsets \(\Delta\hat{\mathbf{Y}_i}\) for the trajectory of agent $i$, and a three-layer MLP \(\phi_\text {refine-conf}\left(\cdot\right)\) is used to generate confidence offsets \(\Delta\hat{\pi}_i\):
\begin{equation}
\Delta\hat{\mathbf{Y}_i}=\phi_{\text {refine-traj}}\left(\boldsymbol{\psi}_i^\text {local},\boldsymbol{\psi}_i^\text {global},\boldsymbol{\psi}_i^{\text {hyper}},\boldsymbol{\psi}_i^\text {consist},\boldsymbol{\psi}_i^{\text {post-int}}\right).\label{eq:17}
\end{equation} 
\begin{equation}
\Delta\hat{\pi}_i=\phi_{\text {refine-conf}}\left(\boldsymbol{\psi}_i^\text {local},\boldsymbol{\psi}_i^\text {global},\boldsymbol{\psi}_i^{\text {hyper}},\boldsymbol{\psi}_i^\text {consist},\boldsymbol{\psi}_i^{\text {post-int}}\right),\label{eq:17}
\end{equation}

With the trajectory offset $\Delta\hat{\mathbf{Y}_i}$ and related confidence offset $\Delta\hat{\pi}_i$, we can compute the final trajectory $\hat{\mathbf{Y}}_{\text{i}}$ and confidence $\hat{\pi}_{\text{i}}$ of agent $i$:
\begin{equation}
\hat{\mathbf{Y}}_{\text{i}} = \hat{\mathbf{Y}}_{\text{base,i}} + \Delta\hat{\mathbf{Y}_i},
\label{eq:18}
\end{equation}
\begin{equation}
\hat{\pi}_{\text{i}} = \hat{\pi}_{\text{base,i}} + \Delta\hat{\pi}_i.
\label{eq:18}
\end{equation}
where $\mathbf{\hat{Y}}_{\text{base},i}$ and $\hat{\pi}_{\text{base,i}}$ are the trajectory proposal and base confidence from the TPN, respectively. As PRN can learn the refinement offsets based on those embeddings. Its refinement offsets will be relatively small if the prediction mode is reasonable; otherwise, the refinement offsets will become very large to correct the unreasonable modes. In this way, we can relocate the trajectories and further approximate the ground truth trajectories. 
\subsection{Three-stage Training Scheme}
We propose a three-stage training scheme to facilitate the learning of interaction features at different levels while preventing TPN and PRN from interfering with CTN during the training process, as shown in Fig. \ref{fig:three-stage}. The three training stages are executed sequentially with each subsequent training
stage reloading weights from the previous stage. In the first stage, we train CTN to achieve coarse trajectory prediction with an auxiliary decoder in an end-to-end manner. In the second stage, we train CTN and
TPN together, and use a main decoder in TPN to generate trajectory proposals. In the third stage, we train
CTN, TPN and PRN together, in which PRN is designed to refine the trajectory proposals to produce the final prediction. \par
Like the previous models \citep{Lanegcn,HiVT,laformer}, we use two types of loss function, including regression loss and classification loss. During the calculation of the regression losses, only the error between the best-predicted trajectory and the ground truth is considered. Therefore, regardless of whether the multimodal trajectories are generated by the decoder or PRN, the rest trajectories with destinations farther from the ground truth destinations will not be used to calculate the regression loss. 
We note that each of the three networks has its corresponding trajectory regression loss and confidence classification loss.\par
In the first stage, the regression loss \(\mathcal{L}_{\text{reg-aux}}\), uses a winner-takes-all strategy to minimize the error between the best-predicted and ground truth trajectories. We employ a negative log-likelihood loss for the auxiliary decoder:
\begin{equation}
\mathcal{L}_{\text{reg-aux}}=-\frac{1}{N} \sum_{t=T_{\text{obs}}+1}^{T_{\text{pred}}} \log \mathrm{P}\left(\mathbf{Y}^t \mid \tilde{\boldsymbol{\mu}}_{k^*}^t, \tilde{\mathbf{b}}_{k^*}^t\right),\label{eq:19}
\end{equation}
where $N$ is the number of predicted time steps, $k^*$ is the best-predicted mode, $\mathbf{Y}^t$ denotes the ground truth trajectory at time step $t$, \(\tilde{\boldsymbol{\mu}}_{k^*}^t\) and \(\tilde{\mathbf{b}}_{k^*}^t\) are parameters of the Laplace distribution of the best-predicted trajectory at time step $t$.
The classification loss \(\mathcal{L}_{\text{cls-aux}}\) is defined with a cross-entropy loss, which optimizes the mixing coefficients to allow the model to discriminate the trajectory confidences of different modes:
\begin{equation}
\mathcal{L}_{\text{cls-aux}}=-\sum_{k=1}^K\pi_k \log \left(\tilde{\pi}_k\right),\label{eq:20}
\end{equation}
where $\pi_k$ denotes the ground truth confidence of mode $k$ and $\tilde{\pi}_k$ denotes the predicted confidence of mode $k$.
Overall, the loss of the first stage $\mathcal{L}_{s1}$ is defined as the sum of the regression loss and the classification loss:
\begin{equation}
\mathcal{L}_{\text{s1}}=\mathcal{L}_{\text{reg-aux}}+\mathcal{L}_{\text{cls-aux}}.\label{eq:21}
\end{equation}
We adhere to the practice of assigning equal weights to the two loss functions, like HiVT \citep{HiVT}.

In the second stage, we introduce the HI module along with a new decoder that serves as the main decoder, which predicts the trajectory proposal $\hat{\mathbf{Y}}_{\text{base}}$ and $\hat{\pi}_{\text{base}}$. We note that the decoder in the first stage serves as the auxiliary decoder to guide the training process, which predicts the coarse trajectory $\tilde{\mathbf{Y}}$ and $\tilde{\pi}$. We reload the weights obtained in the first stage and copy the weights of the auxiliary decoder to the new main decoder. In this stage, we train CTN and TPN together. The loss function in the second stage \(\mathcal{L}_{s2}\) is similar to that in the first stage, defined as:
\begin{equation}
\mathcal{L}_{s2}=\mathcal{L}_{\text {s1}} + \lambda_1(\mathcal{L}_{\text {reg-base}}+\mathcal{L}_{\text {cls-base }}),\label{eq:22}
\end{equation} 
where $\mathcal{L}_{\text{reg-base}}$ and $\mathcal{L}_{\text{cls-base}}$ are the regression loss and the classification loss for the main decoder in TPN, $\lambda_1$ is a hyperparameter used to balance the loss in stage 1 and the loss of the trajectory decoder in TPN. Similarly, $\mathcal{L}_{\text{reg-base}}$ and $\mathcal{L}_{\text{cls-base}}$ is defined with negative
log-likelihood loss and cross-entropy loss:
\begin{equation}
\mathcal{L}_{\text{reg-base}}=-\frac{1}{N} \sum_{t=T_{\text{obs}}+1}^{T_{\text{pred}}} \log \mathrm{P}\left(\mathbf{Y}^t \mid \hat{\boldsymbol{\mu}}_{\text{base},k^*}^t, \hat{\mathbf{b}}_{\text{base}, k^*}^t\right),\label{eq:19}
\end{equation}
\begin{equation}
\mathcal{L}_{\text{cls-base}}=-\sum_{k=1}^K\pi_k \log \left(\hat{\pi}_{\text{base},k}\right),\label{eq:20}
\end{equation}
where \(\tilde{\boldsymbol{\mu}}_{\text{base},k^*}^t\) and \(\tilde{\mathbf{b}}_{\text{base},k^*}^t\) are distribution parameters of the best-predicted trajectory from the trajectory proposals. $\hat{\pi}_
{\text{base},k}$ denotes the predicted confidence in the second stage. 

In the third stage, we introduce PRN to refine the trajectories explicitly. In this stage, we aim to minimize the distance between the refined trajectories and the ground truth trajectories. To this end, we employ the Smooth L1 loss function as regression loss, which is defined as:
\begin{equation}
\mathcal{L}_{\text {reg-refine}}=\frac{1}{N} \sum_{t=T_{\text{obs}} + 1}^{T_{\text{pred}}}\text{SmoothL1}\left(\hat{\mathbf{Y}}_{k^*}^t- \mathbf{Y}^t\right),\label{eq:24}
\end{equation}
\begin{equation}
\operatorname{SmoothL1}(x)= \begin{cases}0.5 x^2 & \text { if }\|x\|<1, \\ \|x\|-0.5 & \text {otherwise}.\end{cases}
\end{equation}
where $\hat{\mathbf{Y}}_{k^*}^t$ denotes the best-predicted trajectory at timestep $t$.
We then use the refined confidences to calculate the classification loss $\mathcal{L}_{\text{cls-refine}}$:
\begin{equation}
\mathcal{L}_{\text{cls-refine}}=-\sum_{k=1}^K\pi_k \log \left(\hat{\pi}_{k}\right).\label{eq:20}
\end{equation}
where $\hat{\pi}_{k}$ denotes the predicted confidence in the third stage.
We train all the networks together in this stage. Thus, the total loss for the third stage \(\mathcal{L}_{s3}\) can be represented as follows:
\begin{equation}
\mathcal{L}_{s3}=\mathcal{L}_{\text {s2}} + 
\lambda_2 (\mathcal{L}_{\text {reg-refine }}+\mathcal{L}_{\text {cls-refine }}),\label{eq:25}
\end{equation}
where $\lambda_2$ is a hyperparameter used to balance the loss of stage 2 and the loss of the trajectory outputs in the refinement network. The overall framework is shown in Fig. \ref{fig:three-stage}. When the three-stage training process is finished, we remove the auxiliary decoder and 
only use the main decoder for model inference.
\subsection{Prediction-based Planning}\label{sec: planning}
\begin{figure}[ht]
    \centering
    \includegraphics[width=0.75\linewidth]{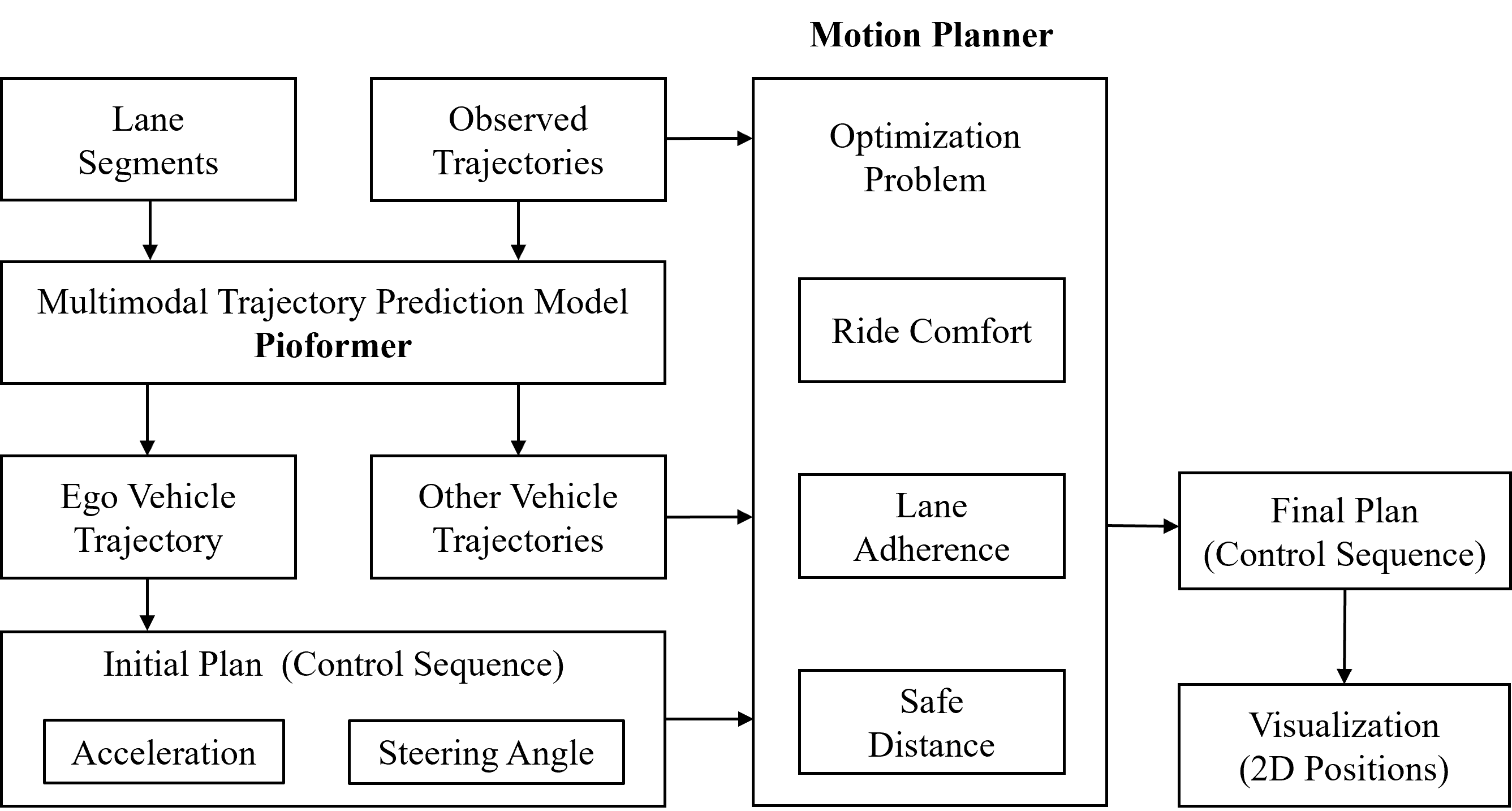}
    \caption{The pipeline of motion planning based on the prediction results. With lane segments and observed trajectory, our model Pioformer can generate the predicted trajectories for ego vehicle and other vehicles. Similar to \citep{planner}, we convert the ego vehicle's predicted trajectory into control sequences, which serve as an initial plan. Then, the initial plan of the ego vehicle, other vehicle-predicted trajectories and observed trajectories are used as the inputs of the motion planner. The motion planner aims to find the optimal control sequence of the ego vehicle by minimizing the cost function, including the ride comfort cost, the lane adherence cost and the safety cost. Finally, we transform the final plan (optimal control) of the ego vehicle back into 2D positions for visualization.}
    \label{fig:planning pipeline}
\end{figure}

Trajectory prediction is usually followed by motion planning, which enables the generation of driving paths that comply with traffic regulations and safety requirements while ensuring ride comfort and traffic efficiency. To demonstrate the use of prediction results from Pioformer, we build a motion planner to solve the optimal control sequence of the ego vehicle based on the predicted trajectories of the ego vehicle and other vehicles, as shown in Fig. \ref{fig:planning pipeline}. With lane segments and observed trajectories, the multimodal trajectory prediction model, Pioformer, can generate the predicted trajectories for the ego vehicle and other vehicles. Similar to \citep{planner}, we convert the ego vehicle's predicted trajectory that has the highest probability into control sequences, which serve as an initial plan for the motion planner. Then, the initial plan of the ego vehicle, other vehicle-predicted trajectories and observed trajectories are used as the inputs of the optimization problem in the motion planner. 
The motion planner aims to find the optimal control sequence of the ego vehicle by minimizing the cost function presented in Fig. \ref{fig:planning pipeline}, including the ride comfort cost, the lane adherence cost and the safety cost. Finally, we transform the final plan (optimal control sequence) of the ego vehicle back into 2D positions for visualization, as shown in Fig. \ref{fig:quant on planning}.  \par
The objective function is defined as the sum of squared residual terms, where each term is a product of a weight $\xi_i$ and cost $\mathcal{C}_i$:
\begin{equation}
\mathbf{U}^*=\arg \min _{\mathbf{U}} \frac{1}{2} \sum_i\left\|\xi_i \mathcal{C}_i\left(\mathbf{U}^i, \hat{\boldsymbol{\kappa}}\right)\right\|^2,
\label{eq:plan}
\end{equation}
where $\mathbf{U}^*$ is the optimal control sequence, which includes acceleration and steering angle. $\hat{\boldsymbol{\kappa}}$ denotes the predicted states of neighboring vehicles, $\mathcal{C}_i$ is a function of  $\mathbf{U}^i \subset \mathbf{U}$ and $\hat{\boldsymbol{\kappa}}$, where $\mathbf{U}=\left\{\mathbf{u}_t\right\}_{t=T_{\text{obs}}+1}^{T_{\text{pred}}}$. The ride comfort cost constrains the acceleration, jerk, steering angle, and steering change rate. The lane adherence cost constrains the distance and angle deviation from the lane center line. The safety cost constrains the safe distance from other traffic participants on the road to avoid collision. For more details, please refer to \citep{planner}.

\begin{table*}[ht]
\centering
\caption{Quantitative results of the comparative study on \textit{Argoverse 1} \citep{argoverse} validation set and the online test set.\dag:The results are obtained using the checkpoints from the official implementation at https://github.com/ZikangZhou/HiVT. Eensemble-based methods and models with parameters over 5000K are in \textbf{bold}. 
}
\resizebox{\linewidth}{!}{
\begin{tabular}{lcccccccc}
\toprule
\multirow{2}{*}{Model} & \multicolumn{3}{c}{Validation  Set} & \multicolumn{4}{c}{Test Set} & \multirow{2}{*}{\#Param  $\downarrow$} \\
 & minADE$_6$ $\downarrow$ & minFDE$_6$ $\downarrow$& MR$_6$ $\downarrow$ & minADE$_6$ $\downarrow$ & minFDE$_6$ $\downarrow$ & MR$_6$ $\downarrow$ & brier-minFDE$_6$ $\downarrow$ &  \\
\midrule
LaneGCN \citep{Lanegcn}  & 0.71 & 1.08 & - & 0.8703 & 1.3622 & 0.1620 & 2.0539& 3701K 	 \\
LaneRCNN \citep{Lanercnn} & 0.77 & 1.19 & 0.08 & 0.9038 & 1.4526 & 0.1232 & 2.1470& -  \\
HOME+GOME \citep{GOHOME}   & - & - & - & 0.8904 & 1.2919 & 0.0846 & 1.8601 & \textbf{Ensemble}	 \\
DenseTNT \citep{DenseTNT}  & 0.73 & 1.05 & 0.10 & 0.8817 & 1.2815 & 0.1258 & - & 1103K \\
mmTransformer \citep{mmtrans}  & 0.72 & 1.21 & 0.09 & 0.8436 & 1.3383 & 0.1540 & 2.0328& 2607K\\
TPCN \citep{TPCN}  & 0.73 & 1.15 & 0.11 & 0.8153 & 1.2442 & 0.1333 & 1.9286 & -	\\
Multipath++ \citep{MultiPath++}  & - & - & - & 0.7897 & 1.2144 & 0.1324 & 1.7932 & \textbf{Ensemble}	 \\
Scene Transformer \citep{scenetrans}  & 0.80 & 1.23 & 0.13 & 0.8026 & 1.2321 & 0.1255 & 1.8868 & \textbf{15296K}	\\
LTP \citep{LTP}  & 0.78 & 1.07 & - & 0.8335 & 1.2955 & 0.1472 & 1.8566 & 1100K	 \\
\rowcolor{gray!15} 
HiVT-64\dag    \citep{HiVT}  &  0.69 & 1.03 & 0.10  & 0.8306 & 1.3053 & 0.1503 & 1.9736 & 662K\\
\rowcolor{gray!15} 
HiVT-128\dag    \citep{HiVT}  & 0.66 & 0.97 & 0.09& 0.7993 & 1.2320 & 0.1369 & 1.9010 &2529K\\
\textbf{Wayformer} \citep{nayakanti2022wayformer} & - & - & - & \textbf{0.7676}& \textbf{1.1616}& \textbf{0.1186}& \textbf{1.7408} & \textbf{Ensemble} \\
FRM \citep{FRM} & 0.68 & 0.99 & - & 0.8165 & 1.2671 & 0.1430 & 1.9365 & - 	 \\
Macformer-S \citep{macformer}  & - & - & - & 0.8490 & 1.2704 & 0.1311 & 1.9021& 879K	 \\
Macformer-L \citep{macformer}  & - & - & - & 0.8188 & 1.2160 & 0.1205 & 1.8275& 2485 K	 \\
\textbf{LAformer} \citep{laformer} & 0.64 & 0.92 & 0.08 & \textbf{0.7720} & \textbf{1.1626} & \textbf{0.1245}  & \textbf{1.8347} & \textbf{2,654K}\\
ProphNet-S \citep{prophnet}  & 0.68 & 0.97 & - & - & - & - & - &-	 \\
\rowcolor{gray!25} 
\textbf{Pioformer (ours)}  & \textbf{0.66} & \textbf{0.95} & \textbf{0.09} & \textbf{0.7939} & \textbf{1.1954} & \textbf{0.1283} & \textbf{1.8608} & \textbf{829K} \\
\bottomrule
\end{tabular}}
\label{tab: sota}
\end{table*}
\begin{table}[width=\linewidth,cols=5,pos=h] 
\centering
\caption{Quantitative results of Pioformer-L on \textit{Argoverse 2} dataset \citep{argoverse2}.}
\begin{tabular*}{\tblwidth}{@{} CCCCC@{} }
\toprule
  Method&  brier-minFDE$_6$ $\downarrow$ &minADE$_6$ $\downarrow$ & minFDE$_6$ $\downarrow$ & MR$_6$ $\downarrow$\\
\midrule
  HiVT-128&  2.51& 0.81& 1.82& 0.27\\
  Piofomer-L&  \textbf{2.25}&\textbf{0.73}& \textbf{1.56}& \textbf{0.22}\\
\bottomrule
\end{tabular*}
\label{tab: argo2}
\end{table}
\begin{table}[width=\linewidth,cols=5,pos=h] 
\centering
\caption{Comparative results in scenarios with three levels of interaction intensity.}
\begin{tabular*}{\tblwidth}{@{} CCCCC@{} }
\toprule   
 Interaction Level & Model & minADE$_6$$\downarrow$ & minFDE$_6$$\downarrow$ & MR$_6$$\downarrow$ \\
\midrule  
\multirow{3}{*}{Slight}& HiVT-64 & 0.51 & 0.68 & 0.04 \\
& HiVT-128 & 0.49 & 0.63 & 0.03 \\
& Pioformer & \textbf{0.49} & \textbf{0.62} & \textbf{0.03} \\
\midrule
\multirow{3}{*}{Moderate}& HiVT-64 & 0.63 & 0.86 & 0.07 \\
& HiVT-128& 0.62 & 0.83 & 0.06\\
& Pioformer & \textbf{0.62} & \textbf{0.83} & \textbf{0.06} \\
\midrule
\multirow{3}{*}{Strong}& HiVT-64 & 0.95 & 1.73 & 0.23\\
& HiVT-128 & 0.90 & 1.60 & 0.21\\
& Pioformer & \textbf{0.88} & \textbf{1.49}& \textbf{0.19}\\
\bottomrule
\end{tabular*}
\label{tab:3 interactions}
\end{table}

\section{Experiments}\label{sec: experiments}
This section introduces the implementation details, including the datasets, evaluation metrics, and baseline methods. We conduct comparative experiments, ablation studies, and sensitivity analysis to evaluate the effectiveness of our model.
\subsection{Experimental Setup} \label{sec:a}
\textbf{Datasets.}
We use \textit{Argoverse 1} Motion Forecasting Dataset to train and validate our motion forecasting model, similar to HiVT \citep{HiVT}. \textit{Argoverse 1} is a public dataset \citep{argoverse}, which includes 324,557 scenarios, each 5 seconds long, including observations for the past 2 seconds and predictions for the future 3 seconds. The most challenging segments such as intersections, vehicles taking left or right turns, and changing lanes are selected.
In \textit{Argoverse 1}, each scenario contains the 2D, birds-eye-view (BEV) centroid of each object sampled at 10 Hz. We note that the BEV data are generated by projecting the 3D perception results from the upstream perception tasks onto a 2D plane.  Except for the trajectory data, \textit{Argoverse 1} also contains HD map information, and the semantic vector maps include lane-level detail, such as lane centerlines, traffic direction, and intersection annotations, which are fed into the proposed prediction model together with the trajectory data. In addition, to further evaluate the generalization ability of our method, we also conduct experiments on benchmark \textit{Argoverse 2} \citep{argoverse2}, which provides trajectory sequences with 5-second observation windows and 6-second prediction horizons. \par
\textbf{Evaluation Metrics.}
We choose the following metrics for evaluations, \textit{i.e.}, Minimum Average Displacement Error ($\min \mathrm{ADE}_K$), Minimum Final Displacement Error ($\min \mathrm{FDE}_K$), Miss Rate ($\mathrm{MR}_K$), and the $\text{brier-}\min\mathrm{FDE}_K$ metric (for more details, see \citep{Lanegcn, DenseTNT, HiVT, huang2023multimodal}). $K$ is the number of prediction modes, and we set $K$ as 6 as requested in the Argoverse Motion Forecasting Competition. 
\begin{itemize}
    \item $\min \mathrm{ADE}_K$ measures the average $l_2$ distances between the best-predicted trajectory and the ground truth. 
    \item $\min \mathrm{FDE}_K$ measures the $l_2$ distance between the endpoint of the best-predicted trajectory and the ground truth. 
    \item $\mathrm{MR}_K$ computes the ratio of scenarios in which the endpoints of the best-predicted trajectory are located more than 2.0 meters from that of the ground truth. 
    \item $\text{brier-}\min\mathrm{FDE}_K$ adds $(1-\hat{\pi})^2$ to the $\min \mathrm{FDE}_K$, where $\hat{\pi}$ corresponds to the probability of the best-predicted trajectory.
\end{itemize}

\textbf{Training Details.}
The hyperparameters of our models are similar to HiVT in \citep{HiVT}. The training batch size is 32, and the initial learning rate is $5 \times 10^{-4}$. The loss weight term $\lambda_1$ is 1, and $\lambda_2$ is 5. The number of nodes $S$ on each hyperedge is set to 4. All experiments are conducted on an RTX 4090 Ti GPU using AdamW optimizer \citep{adamw}. We use the cosine annealing scheduler \citep{SGDR} for 64 epochs at each stage. For more details on network architectures and hyperparameters, see \citep{HiVT}. 
\subsection{Comparative Study}
We compare our method with several state-of-the-art methods regarding the evaluation metrics. To further identify the effectiveness in modeling the interactions, we test our method with the baseline HiVT-64 and HiVT-128 in scenarios with different interaction levels.
\subsubsection{\textbf{Comparison with SOTA}}
The proposed model Pioformer is compared with several state-of-the-art methods regarding the evaluation metrics and the model size (the number of parameters), as shown in Table \ref{tab: sota}. Similarly, we have both a lightweight and a large version of Pioformer,\textit{ i.e.}, Pioformer, and Pioformer-L, which has 64 hidden units and 128 hidden units, respectively. The baseline approaches include HiVT (HiVT-64 and HiVT-128) \citep{HiVT}, the ensemble-based methods such as Wayformer \citep{nayakanti2022wayformer}, multipath++ \citep{MultiPath++}, and HOME+GOME \citep{GOHOME}, and other top-ranking methods on the leaderboard of Argoverse Motion Forecasting Competition over the years. 

The ensemble-based method Wayformer \citep{nayakanti2022wayformer} and LAformer \citep{laformer} have achieved slightly smaller prediction errors than our model, but the two have sacrificed their model sizes and the inference speed, as shown in Table \ref{tab: sota}. The ensemble-based methods often have a model size of more than 5000K, and LAformer \citep{laformer} has a model size of 2654k, three times larger than ours. In contrast, our model size is 829k, and we have achieved a similar level of model accuracy to those large models, \textit{i.e.}, the $\min\mathrm{ADE}_6$ of 0.7939, $\min\mathrm{FDE}_6$ of 1.1954, $\mathrm{MR}_6$ of 0.1283, and $\text{brier-min} \mathrm{FDE}_6$ of 1.8608. In particular, we have smaller prediction errors than the baseline HiVT on both the validation set and test set. Specifically, our model has further reduced the prediction error of HiVT-128 by nearly one-third of its model size. Also, we can achieve much higher accuracy than HiVT-64 without significantly increasing the number of parameters, i.e., 662k vs. 829k. 

We also conduct experiments on another dataset, i.e., \textit{Argoverse 2}, and the results between HiVT-128 and Pioformer-L are shown in Table \ref{tab: argo2}.
Compared with HiVT-128, the prediction errors of Pioformer-L have been substantially reduced regarding all evaluation metrics, and in particular, both $\text{brier-}\min\mathrm{FDE}_6$ and $\min\mathrm{FDE}_6$  decrease by 0.26, showing a very significant improvement in prediction accuracy. As the map encoder of HiVT-128 is not specifically designed to accommodate map information in \textit{Argoverse 2}, the performance of HiVT-128 is suboptimal, and our model shows better performance in  \textit{Argoverse 2}, indicating that our model can generalize better than HiVT. In particular, we note that the parameter size of Pioformer-L is larger than that of HiVT-128, while the small version of Pioformer can achieve a comparable prediction accuracy as the HiVT-128 with one-third of the parameter size. Therefore, we use the lightweight version of Pioformer as our final approach, which can achieve a good trade-off between the model accuracy and model size. 
\par
\subsubsection{\textbf{Comparison in different interaction levels}}
To validate the effectiveness of our model in capturing the high-order interaction features, we test the model in scenarios with three levels of interaction,\textit{ i.e.}, slight interaction, moderate interaction, and strong interaction, as shown in Table \ref{tab:3 interactions}. In scenarios with slight interactions, there are few surrounding vehicles and no turning or lane-changing maneuvers around the central agent. In scenarios with moderate interactions, more vehicles surround the central agent, frequently at intersections, but still without turning maneuvers. In scenarios with strong interactions, the central agent is typically confronted with a dense traffic environment characterized by numerous surrounding vehicles engaging in turning maneuvers. \par
From Table \ref{tab:3 interactions}, the prediction errors increase with the rise of the interaction intensity. Our model has achieved smaller prediction errors than HiVT-64 and HiVT-128 in three interaction scenarios. In strong-interactive scenarios, compared with HiVT-64, there exists a 0.07 decrease in $\min \mathrm{ADE}_6$, 0.24 decrease in $\min \mathrm{FDE}_6$, 0.04 decrease in $\mathrm{MR}_6$. Similarly, in contrast with HiVT-128, we have reduced 0.02 in $\min \mathrm{ADE}_6$, 0.11 in $\min \mathrm{FDE}_6$, and 0.02 in $\mathrm{MR}_6$, respectively. In particular, the gap between our model and HiVT has further enlarged in the strong-interactive scenarios, indicating the strengths of our model in capturing complex interaction behaviors. 
\subsubsection{\textbf{Three-stage training vs. End-to-End training}}
As discussed previously, we execute three training stages sequentially while reloading weights from the previous stage. In stage 1, we train CTN to achieve coarse trajectory prediction. CTN and
TPN are trained together to generate trajectory proposals in stage 2. In stage 3, we train CTN, TPN and PRN together to produce the final prediction. To evaluate the effects of the three-stage training scheme, we compare it with the end-to-end training process, as shown in Table \ref{tab: end2end}. It can be seen that the three-stage training scheme can achieve higher prediction accuracy, i.e., 0.05 decrease in $\min \mathrm{ADE}_6$, 0.1 decrease in $\min \mathrm{FDE}_6$, and 0.01 decrease in $\mathrm{MR}_6$, respectively. Therefore, the proposed three-stage training scheme is beneficial in improving the model's accuracy. 
\begin{figure*}[ht]
    \centering
    \includegraphics[width=\linewidth]{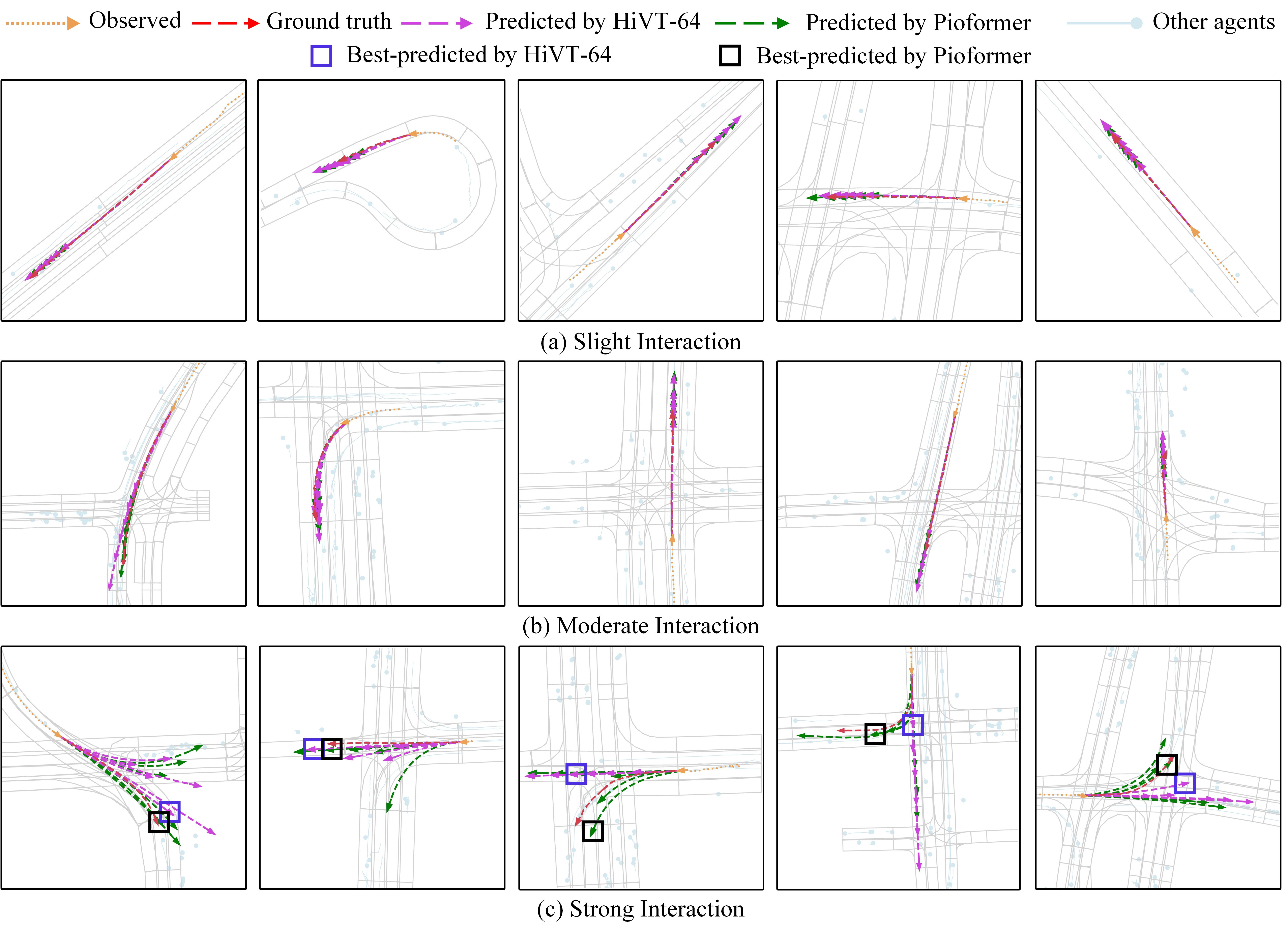}
    \caption{Qualitative comparison of multimodal trajectory predictions of Pioformer and HiVT-64. The best-predicted trajectory is compared with the ground truth trajectory (red) for evaluations. We observe that in scenarios with slight and moderate interactions, both HiVT and Pioformer can generate satisfying predictions, but Pioformer significantly outperforms HiVT-64 in strong-interactive scenarios. The best-predicted trajectory of Pioformer is closer to the ground truth than HiVT-64 (third row), demonstrating the superiority of our method in capturing the complex interactions among agents and the benefits of incorporating the post-interactions into the prediction network. }
    \label{fig: vis. of 3 interactions}
\end{figure*}
\begin{figure*}[ht]
    \centering
    \includegraphics[width=\linewidth]{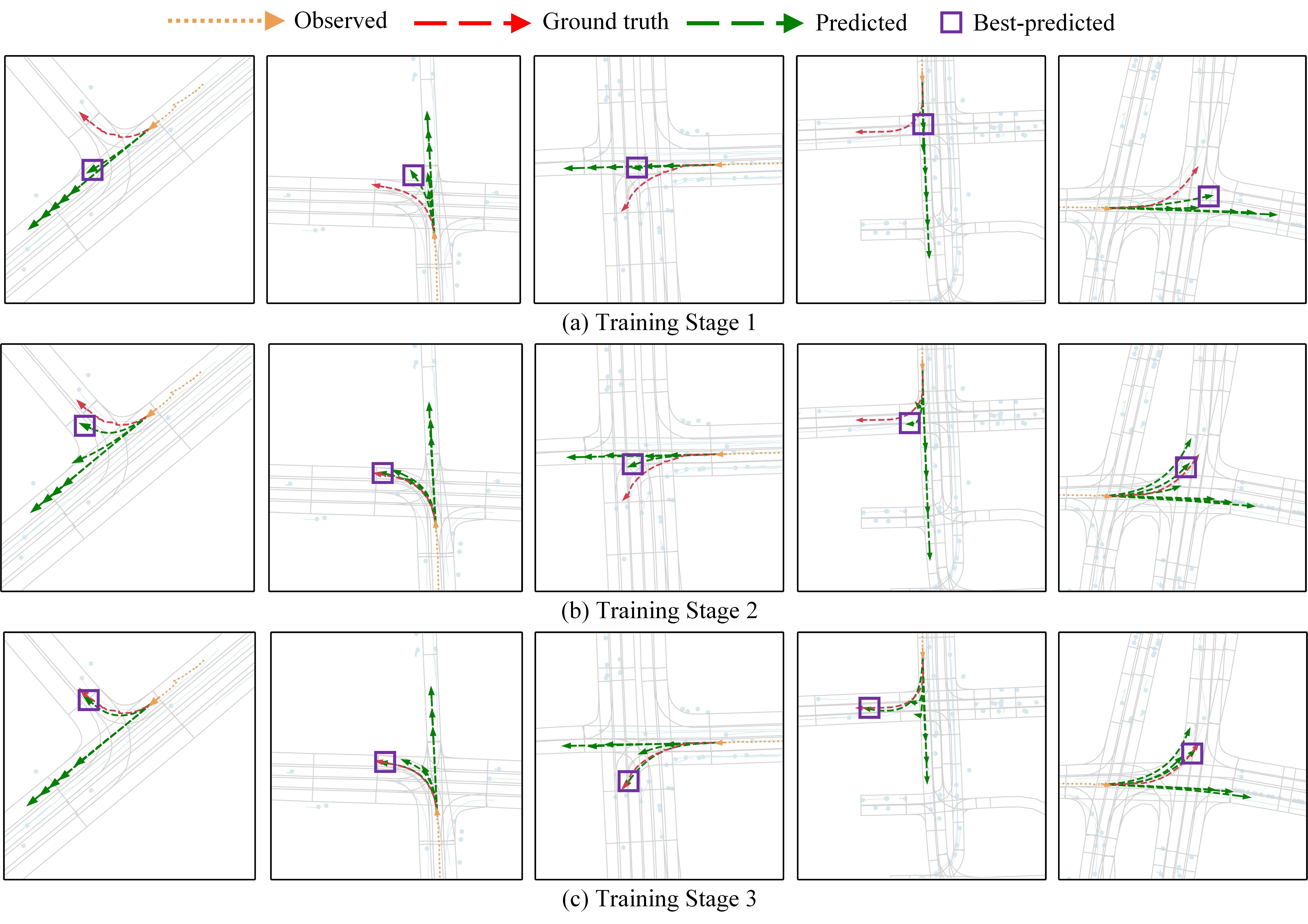}
    \caption{Qualitative comparison of multimodal predictions of Pioformer in three stages. Each column represents a scenario, and we compare the prediction performance of three stages in the same scenario. We observe that as the training stages progress, the predicted trajectories gradually converge toward the ground truths. The final model (training stage 3) demonstrates significantly improved performance compared to the CTN model (training stage 1).}
    \label{fig:quant on three-stage}

\end{figure*}
\begin{table}[width=\linewidth,cols=4,pos=h] 
\centering
\caption{Three-stage training versus end-to-end training.}
    \begin{tabular*}{\tblwidth}{@{} CCCC@{} }
    \toprule 
    Training Scheme& minADE$_6$$\downarrow$& minFDE$_6$ $\downarrow$ & MR$_6$ $\downarrow$ \\
    \midrule 
    End-to-end & 0.71 & 1.05 & 0.10 \\
    Three-stage & \textbf{0.66} & \textbf{0.95} & \textbf{0.09} \\
    \bottomrule
    \end{tabular*}
\label{tab: end2end}
\end{table}
\begin{table}[width=\linewidth,cols=5,pos=h] 
\centering
\caption{Ablation studies on Trajectory Proposal Network and Proposal Refinement Network in our model.}
\begin{tabular*}{\tblwidth}{@{} CCCCC@{} }
\toprule
 Trajectory Proposal Network & Proposal Refinement Network & minADE$_6$$\downarrow$ & minFDE$_6$$\downarrow$ & MR$_6$$\downarrow$ \\
\midrule  
 - & - & 0.69 & 1.03 & 0.10 \\
 $\checkmark$ & - & 0.67 & 0.98 & 0.10 \\
 - & $\checkmark$ & 0.67 & 0.98 & 0.10 \\
 $\checkmark$ & $\checkmark$ & \textbf{0.66} & \textbf{0.95} & \textbf{0.09} \\
\bottomrule
\end{tabular*}
\label{tab: ablation 1}
\end{table}
\begin{table}[width=\linewidth,cols=5,pos=h] 
\caption{Comparison with GNN-based Hyper-Interactor on scenarios of different interaction intensity.}
\begin{tabular*}{\tblwidth}{@{} CCCCC@{} }
\midrule 
Interaction Level & Type of Hyper-Interactor & minADE$_6$$\downarrow$ & minFDE$_6$$\downarrow$ & MR$_6$$\downarrow$ \\
\midrule  
\multirow{2}{*}{Slight}& GNN & 0.50 & 0.65 & 0.03 \\
& HGNN & 0.49 &0.62 &0.03 \\
\midrule
\multirow{2}{*}{Moderate}& GNN & 0.62 & 0.84 & 0.06 \\
& HGNN & 0.62 & 0.83 & 0.06 \\
\midrule
\multirow{2}{*}{Strong}& GNN & 0.92 & 1.64 & 0.22\\
& HGNN & \textbf{0.88} & \textbf{1.49}&\textbf{0.19}\\
\bottomrule
\end{tabular*}
\label{tab:interaction level 2}
\end{table}
\subsection{Ablation Study} \label{sec:c}
We conduct ablation studies to investigate the effect of each part of the model, including the Trajectory Proposal Network (TPN), Proposal Refinement Network (PRN), Hypergraph Neural Network (HGNN), and the three-stage training scheme.   
\subsubsection{\textbf{Trajectory Proposal Network and Proposal Refinement Network}}
We evaluate the effects of TPN and PRN in our model, as shown in Table \ref{tab: ablation 1}. We observe that removing the TPN from our model can increase prediction errors, which indicates that the TPN is beneficial in improving accuracy. Similarly, all evaluation metrics increase when the PRN is removed from our model, indicating the benefits of the PRN. In particular, the prediction errors continue to increase when both TPN and PRN have been removed, i.e., 0.03 increase in $\min \mathrm{ADE}_6$, 0.08 increase in $\min \mathrm{FDE}_6$, and 0.01 increase in $\mathrm{MR}_6$, respectively, which demonstrates the joint benefits of TPN and PRN to our model. 
\subsubsection{\textbf{Effects of HGNN}}
We analyze the effectiveness of the Hypergraph Neural Network (HGNN) in the Hyper-Interactor module of our approach. For comparison, we replace the HGNN with standard GNN, and the prediction results are shown in Table \ref{tab:interaction level 2}. It can be seen that the biggest difference between HGNN and the standard GNN is shown in the strong-interactive scenarios, and HGNN has achieved much smaller prediction errors, i.e., 0.04 reduction in $\min \mathrm{ADE}_6$, 0.15 in $\min\mathrm{FDE}_6$ and 0.03 in $\mathrm{MR}_6$, indicating that HGNN can better capture the complex high-order interactions than the standard GNN in our model. 
\subsubsection{\textbf{Three-stage Training Scheme}}
We compute the number of model parameters and prediction errors in three training stages, as shown in Table \ref{tab: three-stage-process}. It can be seen that the prediction errors decrease from stage 1 to stage 3 without significantly increasing the model size. Specifically, the model learns the underlying patterns of future trajectory prediction in stage 1, which are utilized in the learning of the post-interactions features in stage 2. The previous two stages generate trajectory proposals and interaction features, which are fed as inputs for stage 3, and the model accuracy can be further improved in the final stage. This demonstrates the benefits of the three-stage training scheme in improving the model's accuracy. 
\begin{table}[width=\linewidth,cols=6,pos=h]
\centering
\caption{Ablations on the three-stage training scheme for our model Pioformer on \textit{Argoverse 1}  test set \citep{argoverse}. Best in \textbf{bold}. }
\begin{tabular*}{\tblwidth}{@{} CCCCCC@{} }
\toprule 
Stage & {\#Param}& minADE$_6$ $\downarrow$ & minFDE$_6$ $\downarrow$ & MR$_6$ $\downarrow$ & brier-minFDE$_6$ $\downarrow$ \\
\midrule 
1 & 654K  & 0.8306 & 1.3053 & 0.1503 & 1.9736 \\
2 & 733K & 0.8141 & 1.2683 & 0.1439 & 1.9398 \\
3 & 829K  & \textbf{0.7939} & \textbf{1.1954} & \textbf{0.1283} & \textbf{1.8608} \\
\bottomrule
\end{tabular*}
\label{tab: three-stage-process}
\end{table}
\subsection{Sensitivity Study} \label{sec:d}
The objective of the sensitivity analysis is to evaluate the impact of the loss weights $\lambda_1$ and $\lambda_2$ on the model performance. $\lambda_1$ is used to balance the loss of the auxiliary decoder in the CTN and the loss of the main decoder in the TPN. $\lambda_2$ is used to balance the loss of CTN, TPN, and the loss of the final trajectory outputs in the PRN. We conducted an empirical study to evaluate their impact by varying
their values around the experimental settings, and $\lambda_1$ was varied between 0.5, 1, 5, and $\lambda_2$ was varied between 1, 2, 5, and 10. As shown in Table \ref{tab: lambda}, varying $\lambda_1$ does not impact the model performance, showing that $\lambda_1$ is not a sensitive parameter in our approach. In contrast, the model has the minimum prediction errors in $\min \mathrm{ADE}_6$, and $\min \mathrm{FDE}_6$ when $\lambda_2$ is set as 5, indicating that $\lambda_2$ is a sensitive parameter. An appropriate value of $\lambda_2$ can balance well between the trajectory proposal loss and the refinement loss. 
\begin{table}
    \caption{Sensitivity study of the hyperparameters in the loss functions.}
    \centering
    \begin{subtable}[t]{0.495 \linewidth}
    \vspace{-3.85em}
        \begin{tabular}{ p{1.6cm}<{\centering} p{1.6cm}<{\centering} p{1.6cm}<{\centering} p{1.6cm}<{\centering}}
        \toprule
        $\lambda_1$ & minADE$_6$ $\downarrow$ & minFDE$_6$ $\downarrow$ & MR$_6$ $\downarrow$ \\
        \midrule
        0.5 & 0.67 & 0.98 & 0.10\\
        \rowcolor{gray!25}
        1 & 0.67 & 0.98 & 0.10\\
        5 & 0.67 & 0.98 & 0.10\\
        \bottomrule
        \end{tabular}
    \end{subtable}
    \begin{subtable}[t]{0.495\linewidth}
        \begin{tabular}{ p{1.6cm}<{\centering} p{1.6cm}<{\centering} p{1.6cm}<{\centering} p{1.67cm}<{\centering} }
        \toprule
        $\lambda_2$ & minADE$_6$ $\downarrow$ & minFDE$_6$ $\downarrow$ & MR$_6$ $\downarrow$ \\
        \midrule
        1 & 0.66 & 0.96 & 0.09\\
        2 & 0.66 & 0.96 & 0.09\\
        \rowcolor{gray!25}
        5 & 0.66 & 0.95 & 0.09\\
        10 & 0.67 & 0.95 & 0.09\\
        \bottomrule
        \end{tabular}
    \end{subtable}
    \label{tab: lambda}
\end{table}

\subsection{Qualitative Results}\label{sec:e}
\subsubsection{\textbf{Comparison with HiVT}}
This study focuses on multimodal trajectory prediction, and the prediction results are a distribution of future possible trajectories. We visualize the prediction results of our model Pioformer and the baseline HiVT on different levels of interaction scenarios, as shown in Fig. \ref{fig: vis. of 3 interactions}. Our model Pioformer uses 64 hidden units, and thus, we chose the HiVT-64 for comparison. We observe that both Pioformer and HiVT-64 can generate trajectory predictions that match well with the ground truth trajectories in scenarios with slight and moderate interactions; see Fig. \ref{fig: vis. of 3 interactions} (a) and (b). In contrast, achieving alignments with the ground truth trajectory, especially at intersections, is more challenging. Our model Pioformer has outperformed HiVT-64 in strong-interactive scenarios. The trajectories predicted by Pioformer are located within reasonable ranges, and the best-predicted trajectory (black box) of our model is closer to the ground truth trajectory (red dash line) than that of HiVT-64 (blue box), indicating the superiority of our model in capturing complex interaction behaviors, see Fig. \ref{fig: vis. of 3 interactions} (c).
\subsubsection{\textbf{Three-stage Training Scheme}}
We visualize the prediction results of three training stages, as shown in Fig. \ref{fig:quant on three-stage}. It can be seen that the performance of our model has increased from the stage 1 to the stage 3, with the best-predicted trajectories (marked with box) gradually getting close to the ground truth trajectory (red dash line). In the stage 1, the predicted trajectories often fail to match the correct lanes, resulting in larger displacement errors (see Fig. \ref{fig:quant on three-stage} (a)). In the stage 2, our model implicitly refines the coarse trajectories at the latent space, resulting in a closer approximation to the correct lane and direction (see Fig. \ref{fig:quant on three-stage} (b)). In stage 3, we achieve the final predicted trajectories based on the trajectory proposals from stage 2 and the integrated interaction features. We found that the best-predicted trajectory of the stage 3 has shown the minimum deviations from the ground truth trajectory, as shown in Fig. \ref{fig:quant on three-stage} (c). Overall, the final refinement process at the stage 3 helps bring the deviated trajectories back to the desired lanes and enables more accurate predictions, indicating the effectiveness of the three-stage training scheme. 
\subsubsection{\textbf{Effects on Motion Planning}}  
\begin{figure}[ht]
    \centering
    \includegraphics[width=0.55\linewidth]{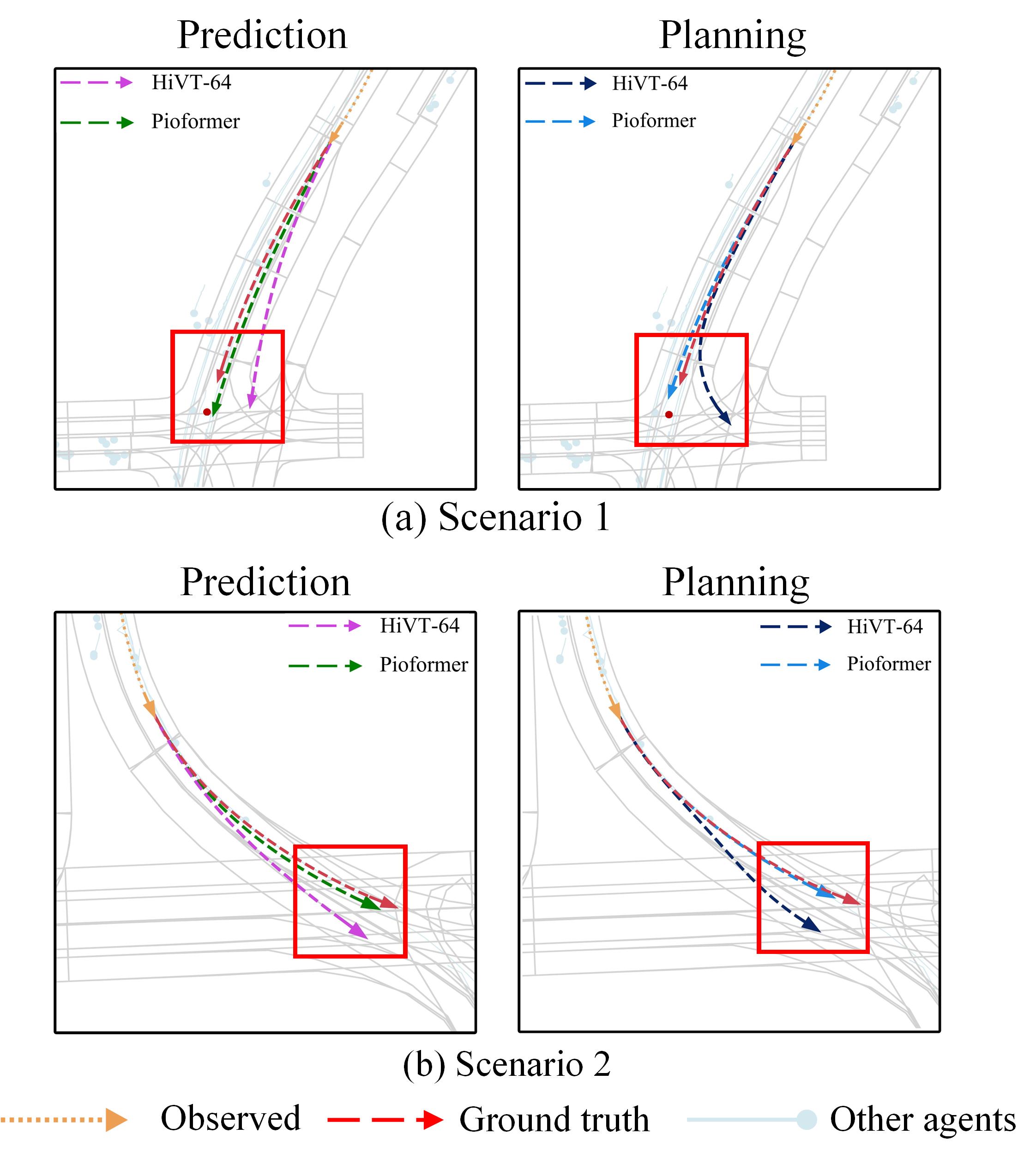}
    \caption{We illustrate how trajectory prediction can benefit motion planning. Prediction results (left) and motion planning based on the predictions  (right). The crucial area is marked with red rectangles, within which the vehicle that is located in front of the ego vehicle is marked in red point. The endpoint of the planned trajectory (right column) based on the predictions from Pioformer can keep a safer distance from the critical vehicle than the predictions. On the left side, the prediction gets very close to the critical vehicle, while the planned trajectory on the right side is far from the vehicle. This shows that trajectory prediction can be used for motion planning and helps avoid critical situations in advance.}
    \label{fig:quant on planning}
\vspace{-1.0em}
\end{figure}

We illustrate how trajectory prediction can benefit motion planning in this study. As discussed in Section \ref{sec: planning}, motion predictions can be employed for motion planning and help improve the performance of motion planning. The prediction results (left column) and the motion planning results (right column) are compared for demonstration, as shown in Fig. \ref{fig:quant on planning}.
The crucial area is marked with red rectangles, within which the most related vehicle that is located right in front of the ego vehicle is marked in red point. It can be seen that our model Pioformer outperforms HiVT-64 in trajectory predictions, resulting in planned trajectories (right column) based on Pioformer being closer to the ground truths than those based on HiVT-64. Besides, the endpoint of the planned trajectory based on predictions from Pioformer can keep a safer distance from the critical vehicle than those in predictions. 
In other words, even if the trajectory endpoint predicted by the Pioformer is close to the critical 
 vehicle, the planned trajectories can maintain a safe distance from it and are closer to the ground truth. This shows that trajectory predictions can bring benefits for motion planning and help planning modules avoid risk situations in advance. Accurately predicted trajectories result in safer planned trajectories that are closer to the ground truth, demonstrating the effectiveness of the prediction-based planning.

\subsection{Discussion}\label{sec: discussion}
This study demonstrates a coarse-to-refine multimodal trajectory prediction model, \textit{i.e.}, Pioformer, which utilizes the high-order post-interaction features for motion predictions. Extensive experiments on \textit{Argoverse 1} indicate that our model can greatly improve prediction accuracy without sacrificing the model size. Our study provides a new insight into the relationship between interaction modeling and motion predictions of multiple agents in the context of intelligent transportation systems. However, the proposed method still needs many improvements for real-world implementations. 
For instance, model latency is essential for the downstream planning task in real-world applications. There exists an inevitable increase in inference latency due to our use of HiVT as the CTN, and the reuse of its GI module in our model. In scenarios with a large number of agents (e.g., 64), the average inference latencies for HiVT-64 and HiVT-128 are 8.4ms and 8.6ms, while Pioformer exhibits a latency of 10.8ms. Despite the slight increase in inference speed, the latency of Pioformer still meets the real-time use cases at 10 Hz. In our future work, we will focus on designing more streamlined network architectures to enhance the inference speed of Pioformer.
In addition, the evaluation metrics $\min \mathrm{ADE}_K$ and $\min \mathrm{FDE}_K$ can cause information leaks when using the best-predicted trajectory for computation \citep{huang2023multimodal}. That is, the trajectory modes that have poor performances (e.g., violations of the traffic rules, large deviations from the ground truth) would be neglected, which is insufficient to provide a comprehensive assessment for the multimodal trajectory models. Instead, the probability-aware metrics can account for the poor modes in the multimodal predictions, which could be used in our future works. For instance, choosing the prediction with the highest probability, or multiple predictions with a probability higher than a threshold for evaluations may provide more useful information. Overall, evaluation metrics are vital in guiding the design of multimodal trajectory prediction models, which should be combined together for a comprehensive and unbiased evaluation. 
\section{Conclusion}\label{sec: Conclusion}
In this study, we propose a post-interactive multimodal trajectory prediction model for autonomous driving, \textit{i.e.}, Pioformer, which can capture the post-interaction features and leverage them to improve the accuracy of the predicted trajectories. Based on the CTN, which serves as a low-order interaction feature extractor and generates coarse trajectories, we propose an HGNN-based TPN to capture the high-order post-interaction features and refine future trajectories in the latent space, resulting in trajectory proposals. To further leverage post-interaction features and explore spatial-temporal consistency, we re-encode the trajectories in the entire horizon and generate both trajectory and confidence offsets to refine the proposals in TPN explicitly. Furthermore, we employ a three-stage training scheme to fully exploit the results of previous training stages, enabling a progressive feature extraction process that promotes stability during the training process. Extensive experiments underscore the superiority of our model to achieve an excellent balance between model size and accuracy. We also demonstrate the effectiveness of the essential components of Pioformer and highlight the necessity of the three-stage training scheme. Moreover, experiment results also indicate that our method can enhance accuracy and reduce the occurrence of unreasonable or hazardous situations in scenarios involving strong interactions. 
\section{Acknowledgments}
We thank Dr. Wenzhao Zheng at UC Berkeley and Dr. Beihao Xia at Huazhong University of Science and Technology for the useful discussion on this study.

\printcredits
\bibliographystyle{cas-model2-names}

\bibliography{cas-refs}

\begin{thebibliography}{62}
\expandafter\ifx\csname natexlab\endcsname\relax\def\natexlab#1{#1}\fi
\providecommand{\url}[1]{\texttt{#1}}
\providecommand{\href}[2]{#2}
\providecommand{\path}[1]{#1}
\providecommand{\DOIprefix}{doi:}
\providecommand{\ArXivprefix}{arXiv:}
\providecommand{\URLprefix}{URL: }
\providecommand{\Pubmedprefix}{pmid:}
\providecommand{\doi}[1]{\href{http://dx.doi.org/#1}{\path{#1}}}
\providecommand{\Pubmed}[1]{\href{pmid:#1}{\path{#1}}}
\providecommand{\bibinfo}[2]{#2}
\ifx\xfnm\relax \def\xfnm[#1]{\unskip,\space#1}\fi
\bibitem[{Alahi et~al.(2016)Alahi, Goel, Ramanathan, Robicquet, Fei-Fei and Savarese}]{Alahi2016}
\bibinfo{author}{Alahi, A.}, \bibinfo{author}{Goel, K.}, \bibinfo{author}{Ramanathan, V.}, \bibinfo{author}{Robicquet, A.}, \bibinfo{author}{Fei-Fei, L.}, \bibinfo{author}{Savarese, S.}, \bibinfo{year}{2016}.
\newblock \bibinfo{title}{Social lstm: Human trajectory prediction in crowded spaces}, in: \bibinfo{booktitle}{Proc. IEEE/CVF Conf. Comput. Vis. Pattern Recognit.}, pp. \bibinfo{pages}{961--971}.
\bibitem[{Cai and Vasconcelos(2018)}]{CascadeRCNN}
\bibinfo{author}{Cai, Z.}, \bibinfo{author}{Vasconcelos, N.}, \bibinfo{year}{2018}.
\newblock \bibinfo{title}{Cascade r-cnn: Delving into high quality object detection}, in: \bibinfo{booktitle}{Proc. IEEE/CVF Conf. Comput. Vis. Pattern Recognit.}
\bibitem[{Chai et~al.(2020)Chai, Sapp, Bansal and Anguelov}]{MultiPath}
\bibinfo{author}{Chai, Y.}, \bibinfo{author}{Sapp, B.}, \bibinfo{author}{Bansal, M.}, \bibinfo{author}{Anguelov, D.}, \bibinfo{year}{2020}.
\newblock \bibinfo{title}{Multipath: Multiple probabilistic anchor trajectory hypotheses for behavior prediction}, in: \bibinfo{editor}{Kaelbling, L.P.}, \bibinfo{editor}{Kragic, D.}, \bibinfo{editor}{Sugiura, K.} (Eds.), \bibinfo{booktitle}{Proc. Conf. Robot Learn.}, \bibinfo{publisher}{PMLR}. pp. \bibinfo{pages}{86--99}.
\bibitem[{Chandra et~al.(2019)Chandra, Bhattacharya, Bera and Manocha}]{Chandra2019}
\bibinfo{author}{Chandra, R.}, \bibinfo{author}{Bhattacharya, U.}, \bibinfo{author}{Bera, A.}, \bibinfo{author}{Manocha, D.}, \bibinfo{year}{2019}.
\newblock \bibinfo{title}{Traphic: Trajectory prediction in dense and heterogeneous traffic using weighted interactions}, in: \bibinfo{booktitle}{Proc. IEEE/CVF Conf. Comput. Vis. Pattern Recognit.}, pp. \bibinfo{pages}{8483--8492}.
\newblock \DOIprefix\doi{10.1109/IEEE/CVF.2019.00868}.
\bibitem[{Chang et~al.(2019)Chang, Lambert, Sangkloy, Singh, Bak, Hartnett, Wang, Carr, Lucey, Ramanan and Hays}]{argoverse}
\bibinfo{author}{Chang, M.F.}, \bibinfo{author}{Lambert, J.}, \bibinfo{author}{Sangkloy, P.}, \bibinfo{author}{Singh, J.}, \bibinfo{author}{Bak, S.}, \bibinfo{author}{Hartnett, A.}, \bibinfo{author}{Wang, D.}, \bibinfo{author}{Carr, P.}, \bibinfo{author}{Lucey, S.}, \bibinfo{author}{Ramanan, D.}, \bibinfo{author}{Hays, J.}, \bibinfo{year}{2019}.
\newblock \bibinfo{title}{Argoverse: 3d tracking and forecasting with rich maps}, in: \bibinfo{booktitle}{Proc. IEEE/CVF Conf. Comput. Vis. Pattern Recognit.}
\bibitem[{Chung et~al.(2014)Chung, Gulcehre, Cho et~al.}]{GRU}
\bibinfo{author}{Chung, J.}, \bibinfo{author}{Gulcehre, C.}, \bibinfo{author}{Cho, K.}, et~al., \bibinfo{year}{2014}.
\newblock \bibinfo{title}{Empirical evaluation of gated recurrent neural networks on sequence modeling}.
\newblock \bibinfo{journal}{ArXiv:1412.3555} .
\bibitem[{Ding et~al.(2020)Ding, Wang, Li, Li and Liu}]{hgnn_nlp}
\bibinfo{author}{Ding, K.}, \bibinfo{author}{Wang, J.}, \bibinfo{author}{Li, J.}, \bibinfo{author}{Li, D.}, \bibinfo{author}{Liu, H.}, \bibinfo{year}{2020}.
\newblock \bibinfo{title}{Be more with less: Hypergraph attention networks for inductive text classification}, in: \bibinfo{booktitle}{Proc. Conf. Empir. Methods Nat. Lang. Process.}, pp. \bibinfo{pages}{4927--4936}.
\newblock \DOIprefix\doi{10.18653/v1/2020.emnlp-main.399}.
\bibitem[{Fang et~al.(2020)Fang, Jiang, Shi and Zhou}]{Fang2020TPNetTP}
\bibinfo{author}{Fang, L.}, \bibinfo{author}{Jiang, Q.}, \bibinfo{author}{Shi, J.}, \bibinfo{author}{Zhou, B.}, \bibinfo{year}{2020}.
\newblock \bibinfo{title}{Tpnet: Trajectory proposal network for motion prediction}.
\newblock \bibinfo{journal}{Proc. IEEE/CVF Conf. Comput. Vis. Pattern Recognit.} , \bibinfo{pages}{6796--6805}.
\bibitem[{Feng et~al.(2023)Feng, Zhou, Lin, Zhang, Xu, Zhang, Zhou and Shen}]{macformer}
\bibinfo{author}{Feng, C.}, \bibinfo{author}{Zhou, H.}, \bibinfo{author}{Lin, H.}, \bibinfo{author}{Zhang, Z.}, \bibinfo{author}{Xu, Z.}, \bibinfo{author}{Zhang, C.}, \bibinfo{author}{Zhou, B.}, \bibinfo{author}{Shen, S.}, \bibinfo{year}{2023}.
\newblock \bibinfo{title}{Macformer: Map-agent coupled transformer for real-time and robust trajectory prediction}.
\newblock \bibinfo{journal}{{IEEE} Robot. Autom. Lett.} \bibinfo{volume}{8}, \bibinfo{pages}{6795--6802}.
\newblock \DOIprefix\doi{10.1109/LRA.2023.3311351}.
\bibitem[{Feng et~al.(2018)Feng, You, Zhang, Ji and Gao}]{HGNN}
\bibinfo{author}{Feng, Y.}, \bibinfo{author}{You, H.}, \bibinfo{author}{Zhang, Z.}, \bibinfo{author}{Ji, R.}, \bibinfo{author}{Gao, Y.}, \bibinfo{year}{2018}.
\newblock \bibinfo{title}{Hypergraph neural networks}, in: \bibinfo{booktitle}{Proc. AAAI Conf. Artif. Intell.}
\bibitem[{Gao et~al.(2020)Gao, Sun, Zhao, Shen, Anguelov, Li and Schmid}]{gao2020vectornet}
\bibinfo{author}{Gao, J.}, \bibinfo{author}{Sun, C.}, \bibinfo{author}{Zhao, H.}, \bibinfo{author}{Shen, Y.}, \bibinfo{author}{Anguelov, D.}, \bibinfo{author}{Li, C.}, \bibinfo{author}{Schmid, C.}, \bibinfo{year}{2020}.
\newblock \bibinfo{title}{Vectornet: Encoding hd maps and agent dynamics from vectorized representation}, in: \bibinfo{booktitle}{Proc. IEEE/CVF Conf. Comput. Vis. Pattern Recognit.}, pp. \bibinfo{pages}{11525--11533}.
\bibitem[{Gao et~al.(2023)Gao, Feng, Ji and Ji}]{HGNN+}
\bibinfo{author}{Gao, Y.}, \bibinfo{author}{Feng, Y.}, \bibinfo{author}{Ji, S.}, \bibinfo{author}{Ji, R.}, \bibinfo{year}{2023}.
\newblock \bibinfo{title}{Hgnn+: General hypergraph neural networks}.
\newblock \bibinfo{journal}{IEEE Trans. Pattern Anal. Mach. Intell.} \bibinfo{volume}{45}, \bibinfo{pages}{3181--3199}.
\newblock \DOIprefix\doi{10.1109/TPAMI.2022.3182052}.
\bibitem[{Gilles et~al.(2022)Gilles, Sabatini, Tsishkou, Stanciulescu and Moutarde}]{GOHOME}
\bibinfo{author}{Gilles, T.}, \bibinfo{author}{Sabatini, S.}, \bibinfo{author}{Tsishkou, D.}, \bibinfo{author}{Stanciulescu, B.}, \bibinfo{author}{Moutarde, F.}, \bibinfo{year}{2022}.
\newblock \bibinfo{title}{Gohome: Graph-oriented heatmap output for future motion estimation}, in: \bibinfo{booktitle}{IEEE Int. Conf. Robot. Autom.}, pp. \bibinfo{pages}{9107--9114}.
\newblock \DOIprefix\doi{10.1109/ICRA46639.2022.9812253}.
\bibitem[{Goodfellow et~al.(2014)Goodfellow, Pouget-Abadie, Mirza, Xu, Warde-Farley, Ozair, Courville and Bengio}]{GAN}
\bibinfo{author}{Goodfellow, I.}, \bibinfo{author}{Pouget-Abadie, J.}, \bibinfo{author}{Mirza, M.}, \bibinfo{author}{Xu, B.}, \bibinfo{author}{Warde-Farley, D.}, \bibinfo{author}{Ozair, S.}, \bibinfo{author}{Courville, A.}, \bibinfo{author}{Bengio, Y.}, \bibinfo{year}{2014}.
\newblock \bibinfo{title}{Generative adversarial nets}, in: \bibinfo{editor}{Ghahramani, Z.}, \bibinfo{editor}{Welling, M.}, \bibinfo{editor}{Cortes, C.}, \bibinfo{editor}{Lawrence, N.}, \bibinfo{editor}{Weinberger, K.} (Eds.), \bibinfo{booktitle}{Adv. Neural Inf. Process. Syst.}, \bibinfo{publisher}{Curran Associates, Inc.}
\bibitem[{Gu et~al.(2021)Gu, Sun and Zhao}]{DenseTNT}
\bibinfo{author}{Gu, J.}, \bibinfo{author}{Sun, C.}, \bibinfo{author}{Zhao, H.}, \bibinfo{year}{2021}.
\newblock \bibinfo{title}{Densetnt: End-to-end trajectory prediction from dense goal sets}, in: \bibinfo{booktitle}{Proc. IEEE/CVF Int. Conf. Comput. Vis.}, pp. \bibinfo{pages}{15283--15292}.
\newblock \DOIprefix\doi{10.1109/ICCV48922.2021.01502}.
\bibitem[{Gu et~al.(2022)Gu, Chen, Li, Lin, Rao, Zhou and Lu}]{gu2022stochastic}
\bibinfo{author}{Gu, T.}, \bibinfo{author}{Chen, G.}, \bibinfo{author}{Li, J.}, \bibinfo{author}{Lin, C.}, \bibinfo{author}{Rao, Y.}, \bibinfo{author}{Zhou, J.}, \bibinfo{author}{Lu, J.}, \bibinfo{year}{2022}.
\newblock \bibinfo{title}{Stochastic trajectory prediction via motion indeterminacy diffusion}, in: \bibinfo{booktitle}{Proc. IEEE/CVF Conf. Comput. Vis. Pattern Recognit.}, pp. \bibinfo{pages}{17113--17122}.
\bibitem[{Gupta et~al.(2018a)Gupta, Johnson, Fei-Fei, Savarese and Alahi}]{gupta2018social}
\bibinfo{author}{Gupta, A.}, \bibinfo{author}{Johnson, J.}, \bibinfo{author}{Fei-Fei, L.}, \bibinfo{author}{Savarese, S.}, \bibinfo{author}{Alahi, A.}, \bibinfo{year}{2018}a.
\newblock \bibinfo{title}{Social gan: Socially acceptable trajectories with generative adversarial networks}, in: \bibinfo{booktitle}{Proc. IEEE/CVF Conf. Comput. Vis. Pattern Recognit.}, pp. \bibinfo{pages}{2255--2264}.
\bibitem[{Gupta et~al.(2018b)Gupta, Johnson, Fei-Fei, Savarese and Alahi}]{socialgan}
\bibinfo{author}{Gupta, A.}, \bibinfo{author}{Johnson, J.}, \bibinfo{author}{Fei-Fei, L.}, \bibinfo{author}{Savarese, S.}, \bibinfo{author}{Alahi, A.}, \bibinfo{year}{2018}b.
\newblock \bibinfo{title}{Social gan: Socially acceptable trajectories with generative adversarial networks}, in: \bibinfo{booktitle}{Proc. IEEE/CVF Conf. Comput. Vis. Pattern Recognit.}, pp. \bibinfo{pages}{2255--2264}.
\newblock \DOIprefix\doi{10.1109/CVPR.2018.00240}.
\bibitem[{Han et~al.(2023)Han, Wang, Kundu, Ding and Wang}]{hgnn_cv}
\bibinfo{author}{Han, Y.}, \bibinfo{author}{Wang, P.}, \bibinfo{author}{Kundu, S.}, \bibinfo{author}{Ding, Y.}, \bibinfo{author}{Wang, Z.}, \bibinfo{year}{2023}.
\newblock \bibinfo{title}{Vision hgnn: An image is more than a graph of nodes}, in: \bibinfo{booktitle}{Proc. IEEE/CVF Int. Conf. Comput. Vis.}, pp. \bibinfo{pages}{19878--19888}.
\bibitem[{Helbing and Molnar(1995)}]{helbing1995}
\bibinfo{author}{Helbing, D.}, \bibinfo{author}{Molnar, P.}, \bibinfo{year}{1995}.
\newblock \bibinfo{title}{Social force model for pedestrian dynamics}.
\newblock \bibinfo{journal}{Phys. Rev. E} \bibinfo{volume}{51}, \bibinfo{pages}{4282}.
\bibitem[{Ho et~al.(2020)Ho, Jain and Abbeel}]{DDPM}
\bibinfo{author}{Ho, J.}, \bibinfo{author}{Jain, A.}, \bibinfo{author}{Abbeel, P.}, \bibinfo{year}{2020}.
\newblock \bibinfo{title}{Denoising diffusion probabilistic models}, in: \bibinfo{editor}{Larochelle, H.}, \bibinfo{editor}{Ranzato, M.}, \bibinfo{editor}{Hadsell, R.}, \bibinfo{editor}{Balcan, M.}, \bibinfo{editor}{Lin, H.} (Eds.), \bibinfo{booktitle}{Adv. Neural Inf. Process. Syst.}, \bibinfo{publisher}{Curran Associates, Inc.}. pp. \bibinfo{pages}{6840--6851}.
\bibitem[{Huang et~al.(2023a)Huang, Xue, Pagnucco, Salim and Song}]{huang2023multimodal}
\bibinfo{author}{Huang, R.}, \bibinfo{author}{Xue, H.}, \bibinfo{author}{Pagnucco, M.}, \bibinfo{author}{Salim, F.}, \bibinfo{author}{Song, Y.}, \bibinfo{year}{2023}a.
\newblock \bibinfo{title}{Multimodal trajectory prediction: A survey}.
\newblock \bibinfo{journal}{ArXiv:2302.10463} .
\bibitem[{Huang et~al.(2022)Huang, Du, Yang, Zhou, Zhang and Chen}]{TIV_survey}
\bibinfo{author}{Huang, Y.}, \bibinfo{author}{Du, J.}, \bibinfo{author}{Yang, Z.}, \bibinfo{author}{Zhou, Z.}, \bibinfo{author}{Zhang, L.}, \bibinfo{author}{Chen, H.}, \bibinfo{year}{2022}.
\newblock \bibinfo{title}{A survey on trajectory-prediction methods for autonomous driving}.
\newblock \bibinfo{journal}{{IEEE} Trans. Intell. Veh.} \bibinfo{volume}{7}, \bibinfo{pages}{652--674}.
\newblock \DOIprefix\doi{10.1109/TIV.2022.3167103}.
\bibitem[{Huang et~al.(2023b)Huang, Liu, Wu and Lv}]{planner}
\bibinfo{author}{Huang, Z.}, \bibinfo{author}{Liu, H.}, \bibinfo{author}{Wu, J.}, \bibinfo{author}{Lv, C.}, \bibinfo{year}{2023}b.
\newblock \bibinfo{title}{Differentiable integrated motion prediction and planning with learnable cost function for autonomous driving}.
\newblock \bibinfo{journal}{IEEE Trans. Neural Networks Learn. Syst.} , \bibinfo{pages}{1--15}\DOIprefix\doi{10.1109/TNNLS.2023.3283542}.
\bibitem[{{Ilya Loshchilov and Frank Hutter}(2017)}]{SGDR}
\bibinfo{author}{{Ilya Loshchilov and Frank Hutter}}, \bibinfo{year}{2017}.
\newblock \bibinfo{title}{Sgdr: Stochastic gradient descent with warm restarts}.
\newblock \bibinfo{journal}{ArXiv:1608.03983} \href{http://arxiv.org/abs/1608.03983}{\tt arXiv:1608.03983}.
\bibitem[{Jia et~al.(2023)Jia, Wu, Chen, Liu, Li and Yan}]{HDGT}
\bibinfo{author}{Jia, X.}, \bibinfo{author}{Wu, P.}, \bibinfo{author}{Chen, L.}, \bibinfo{author}{Liu, Y.}, \bibinfo{author}{Li, H.}, \bibinfo{author}{Yan, J.}, \bibinfo{year}{2023}.
\newblock \bibinfo{title}{Hdgt: Heterogeneous driving graph transformer for multi-agent trajectory prediction via scene encoding}.
\newblock \bibinfo{journal}{IEEE Trans. Pattern Anal. Mach. Intell.} .
\bibitem[{Jiang et~al.(2019)Jiang, Zhang, Lin, Tang and Luo}]{GCN}
\bibinfo{author}{Jiang, B.}, \bibinfo{author}{Zhang, Z.}, \bibinfo{author}{Lin, D.}, \bibinfo{author}{Tang, J.}, \bibinfo{author}{Luo, B.}, \bibinfo{year}{2019}.
\newblock \bibinfo{title}{Semi-supervised learning with graph learning-convolutional networks}, in: \bibinfo{booktitle}{Proc. IEEE/CVF Conf. Comput. Vis. Pattern Recognit.}, pp. \bibinfo{pages}{11305--11312}.
\newblock \DOIprefix\doi{10.1109/CVPR.2019.01157}.
\bibitem[{Jiao et~al.(2023)Jiao, Wang, Liu, Huang and Zhu}]{kinematicsaware}
\bibinfo{author}{Jiao, R.}, \bibinfo{author}{Wang, Y.}, \bibinfo{author}{Liu, X.}, \bibinfo{author}{Huang, C.}, \bibinfo{author}{Zhu, Q.}, \bibinfo{year}{2023}.
\newblock \bibinfo{title}{Kinematics-aware trajectory generation and prediction with latent stochastic differential modeling}.
\newblock \bibinfo{journal}{ArXiv:2309.09317} .
\bibitem[{Kim et~al.(2020)Kim, Kang, On, Heo and Zhang}]{hgnn_multi}
\bibinfo{author}{Kim, E.S.}, \bibinfo{author}{Kang, W.Y.}, \bibinfo{author}{On, K.W.}, \bibinfo{author}{Heo, Y.J.}, \bibinfo{author}{Zhang, B.T.}, \bibinfo{year}{2020}.
\newblock \bibinfo{title}{Hypergraph attention networks for multimodal learning}, in: \bibinfo{booktitle}{Proc. IEEE/CVF Conf. Comput. Vis. Pattern Recognit.}, pp. \bibinfo{pages}{14569--14578}.
\newblock \DOIprefix\doi{10.1109/CVPR42600.2020.01459}.
\bibitem[{Kossale et~al.(2022)Kossale, Airaj and Darouichi}]{mode_collapse}
\bibinfo{author}{Kossale, Y.}, \bibinfo{author}{Airaj, M.}, \bibinfo{author}{Darouichi, A.}, \bibinfo{year}{2022}.
\newblock \bibinfo{title}{Mode collapse in generative adversarial networks: An overview}, in: \bibinfo{booktitle}{Int. Conf. Optim. Appl.}, pp. \bibinfo{pages}{1--6}.
\newblock \DOIprefix\doi{10.1109/ICOA55659.2022.9934291}.
\bibitem[{Lee et~al.(2017)Lee, Choi, Vernaza, Choy, Torr and Chandraker}]{DESIRE}
\bibinfo{author}{Lee, N.}, \bibinfo{author}{Choi, W.}, \bibinfo{author}{Vernaza, P.}, \bibinfo{author}{Choy, C.B.}, \bibinfo{author}{Torr, P.H.S.}, \bibinfo{author}{Chandraker, M.}, \bibinfo{year}{2017}.
\newblock \bibinfo{title}{Desire: Distant future prediction in dynamic scenes with interacting agents}, in: \bibinfo{booktitle}{Proc. IEEE/CVF Conf. Comput. Vis. Pattern Recognit.}, pp. \bibinfo{pages}{2165--2174}.
\newblock \DOIprefix\doi{10.1109/CVPR.2017.233}.
\bibitem[{Li et~al.(2023)Li, Zhang, Xia, Zheng, Zhang, Yi, Jin and Zhao}]{PIH}
\bibinfo{author}{Li, D.}, \bibinfo{author}{Zhang, Q.}, \bibinfo{author}{Xia, Z.}, \bibinfo{author}{Zheng, Y.}, \bibinfo{author}{Zhang, K.}, \bibinfo{author}{Yi, M.}, \bibinfo{author}{Jin, W.}, \bibinfo{author}{Zhao, D.}, \bibinfo{year}{2023}.
\newblock \bibinfo{title}{Planning-inspired hierarchical trajectory prediction via lateral-longitudinal decomposition for autonomous driving}.
\newblock \bibinfo{journal}{{IEEE} Trans. Intell. Veh.} , \bibinfo{pages}{1--12}\DOIprefix\doi{10.1109/TIV.2023.3307116}.
\bibitem[{Li et~al.(2020)Li, Lu, Wang and Li}]{Pedestrian}
\bibinfo{author}{Li, Y.}, \bibinfo{author}{Lu, X.Y.}, \bibinfo{author}{Wang, J.}, \bibinfo{author}{Li, K.}, \bibinfo{year}{2020}.
\newblock \bibinfo{title}{Pedestrian trajectory prediction combining probabilistic reasoning and sequence learning}.
\newblock \bibinfo{journal}{{IEEE} Trans. Intell. Veh.} \bibinfo{volume}{5}, \bibinfo{pages}{461--474}.
\newblock \DOIprefix\doi{10.1109/TIV.2020.2966117}.
\bibitem[{Liang et~al.(2020)Liang, Yang, Hu, Chen, Liao, Feng and Urtasun}]{Lanegcn}
\bibinfo{author}{Liang, M.}, \bibinfo{author}{Yang, B.}, \bibinfo{author}{Hu, R.}, \bibinfo{author}{Chen, Y.}, \bibinfo{author}{Liao, R.}, \bibinfo{author}{Feng, S.}, \bibinfo{author}{Urtasun, R.}, \bibinfo{year}{2020}.
\newblock \bibinfo{title}{Learning lane graph representations for motion forecasting}, in: \bibinfo{booktitle}{Proc. Eur. Conf. Comput. Vis.}, pp. \bibinfo{pages}{541--556}.
\bibitem[{Liang et~al.(2023)Liang, Li, Zhou and Li}]{STglow}
\bibinfo{author}{Liang, R.}, \bibinfo{author}{Li, Y.}, \bibinfo{author}{Zhou, J.}, \bibinfo{author}{Li, X.}, \bibinfo{year}{2023}.
\newblock \bibinfo{title}{Stglow: A flow-based generative framework with dual-graphormer for pedestrian trajectory prediction}.
\newblock \bibinfo{journal}{IEEE Trans. Neural Networks Learn. Syst.} , \bibinfo{pages}{1--14}\DOIprefix\doi{10.1109/TNNLS.2023.3294998}.
\bibitem[{Lin et~al.(2020)Lin, Liu, Milan, Shen and Reid}]{refinenet}
\bibinfo{author}{Lin, G.}, \bibinfo{author}{Liu, F.}, \bibinfo{author}{Milan, A.}, \bibinfo{author}{Shen, C.}, \bibinfo{author}{Reid, I.}, \bibinfo{year}{2020}.
\newblock \bibinfo{title}{Refinenet: Multi-path refinement networks for dense prediction}.
\newblock \bibinfo{journal}{IEEE Trans. Pattern Anal. Mach. Intell.} \bibinfo{volume}{42}, \bibinfo{pages}{1228--1242}.
\newblock \DOIprefix\doi{10.1109/TPAMI.2019.2893630}.
\bibitem[{Liu et~al.(2024)Liu, Cheng, Chen, Broszio, Li, Zhao, Sester and Yang}]{laformer}
\bibinfo{author}{Liu, M.}, \bibinfo{author}{Cheng, H.}, \bibinfo{author}{Chen, L.}, \bibinfo{author}{Broszio, H.}, \bibinfo{author}{Li, J.}, \bibinfo{author}{Zhao, R.}, \bibinfo{author}{Sester, M.}, \bibinfo{author}{Yang, M.Y.}, \bibinfo{year}{2024}.
\newblock \bibinfo{title}{Laformer: Trajectory prediction for autonomous driving with lane-aware scene constraints}, in: \bibinfo{booktitle}{Proc. IEEE/CVF Conf. Comput. Vis. Pattern Recognit.}, pp. \bibinfo{pages}{2039--2049}.
\bibitem[{Liu et~al.(2021)Liu, Zhang, Fang, Jiang and Zhou}]{mmtrans}
\bibinfo{author}{Liu, Y.}, \bibinfo{author}{Zhang, J.}, \bibinfo{author}{Fang, L.}, \bibinfo{author}{Jiang, Q.}, \bibinfo{author}{Zhou, B.}, \bibinfo{year}{2021}.
\newblock \bibinfo{title}{Multimodal motion prediction with stacked transformers}, in: \bibinfo{booktitle}{Proc. IEEE/CVF Conf. Comput. Vis. Pattern Recognit.}, pp. \bibinfo{pages}{7573--7582}.
\newblock \DOIprefix\doi{10.1109/CVPR46437.2021.00749}.
\bibitem[{Loshchilov and Hutter(2019)}]{adamw}
\bibinfo{author}{Loshchilov, I.}, \bibinfo{author}{Hutter, F.}, \bibinfo{year}{2019}.
\newblock \bibinfo{title}{Decoupled weight decay regularization}, in: \bibinfo{booktitle}{Proc. Int. Conf. Learn. Represent.}
\bibitem[{Luo et~al.(2023)Luo, E, Yang, Guo, Sun, Yao, Tang, Wan, Song and Lin}]{hgnn_dm}
\bibinfo{author}{Luo, H.}, \bibinfo{author}{E, H.}, \bibinfo{author}{Yang, Y.}, \bibinfo{author}{Guo, Y.}, \bibinfo{author}{Sun, M.}, \bibinfo{author}{Yao, T.}, \bibinfo{author}{Tang, Z.}, \bibinfo{author}{Wan, K.}, \bibinfo{author}{Song, M.}, \bibinfo{author}{Lin, W.}, \bibinfo{year}{2023}.
\newblock \bibinfo{title}{{HAHE}: Hierarchical attention for hyper-relational knowledge graphs in global and local level}, in: \bibinfo{booktitle}{Proc. Annu. Meet. Assoc. Comput Linguist.}, pp. \bibinfo{pages}{8095--8107}.
\newblock \DOIprefix\doi{10.18653/v1/2023.acl-long.450}.
\bibitem[{Ma et~al.(2019)Ma, Zhu, Zhang, Yang, Wang and Manocha}]{ma2019trafficpredict}
\bibinfo{author}{Ma, Y.}, \bibinfo{author}{Zhu, X.}, \bibinfo{author}{Zhang, S.}, \bibinfo{author}{Yang, R.}, \bibinfo{author}{Wang, W.}, \bibinfo{author}{Manocha, D.}, \bibinfo{year}{2019}.
\newblock \bibinfo{title}{Trafficpredict: Trajectory prediction for heterogeneous traffic-agents}, in: \bibinfo{booktitle}{Proc. AAAI Conf. Artif. Intell.}, pp. \bibinfo{pages}{6120--6127}.
\bibitem[{Mohamed et~al.(2020)Mohamed, Qian, Elhoseiny and Claudel}]{mohamed2020}
\bibinfo{author}{Mohamed, A.}, \bibinfo{author}{Qian, K.}, \bibinfo{author}{Elhoseiny, M.}, \bibinfo{author}{Claudel, C.}, \bibinfo{year}{2020}.
\newblock \bibinfo{title}{Social-stgcnn: A social spatio-temporal graph convolutional neural network for human trajectory prediction}, in: \bibinfo{booktitle}{Proc. IEEE/CVF Conf. Comput. Vis. Pattern Recognit.}, pp. \bibinfo{pages}{14424--14432}.
\bibitem[{Nayakanti et~al.(2022)Nayakanti, Al-Rfou, Zhou, Goel, Refaat and Sapp}]{nayakanti2022wayformer}
\bibinfo{author}{Nayakanti, N.}, \bibinfo{author}{Al-Rfou, R.}, \bibinfo{author}{Zhou, A.}, \bibinfo{author}{Goel, K.}, \bibinfo{author}{Refaat, K.S.}, \bibinfo{author}{Sapp, B.}, \bibinfo{year}{2022}.
\newblock \bibinfo{title}{Wayformer: Motion forecasting via simple \& efficient attention networks}.
\newblock \bibinfo{journal}{ArXiv:2207.05844} \href{http://arxiv.org/abs/2207.05844}{\tt arXiv:2207.05844}.
\bibitem[{Ngiam et~al.(2022)Ngiam, Caine, Vasudevan, Zhang, Chiang, Ling, Roelofs, Bewley, Liu, Venugopal, Weiss, Sapp, Chen and Shlens}]{scenetrans}
\bibinfo{author}{Ngiam, J.}, \bibinfo{author}{Caine, B.}, \bibinfo{author}{Vasudevan, V.}, \bibinfo{author}{Zhang, Z.}, \bibinfo{author}{Chiang, H.T.L.}, \bibinfo{author}{Ling, J.}, \bibinfo{author}{Roelofs, R.}, \bibinfo{author}{Bewley, A.}, \bibinfo{author}{Liu, C.}, \bibinfo{author}{Venugopal, A.}, \bibinfo{author}{Weiss, D.}, \bibinfo{author}{Sapp, B.}, \bibinfo{author}{Chen, Z.}, \bibinfo{author}{Shlens, J.}, \bibinfo{year}{2022}.
\newblock \bibinfo{title}{Scene transformer: A unified architecture for predicting multiple agent trajectories}.
\newblock \bibinfo{journal}{ArXiv:2106.08417} \href{http://arxiv.org/abs/2106.08417}{\tt arXiv:2106.08417}.
\bibitem[{Papamakarios et~al.(2019)Papamakarios, Nalisnick, Rezende, Mohamed and Lakshminarayanan}]{NF}
\bibinfo{author}{Papamakarios, G.}, \bibinfo{author}{Nalisnick, E.T.}, \bibinfo{author}{Rezende, D.J.}, \bibinfo{author}{Mohamed, S.}, \bibinfo{author}{Lakshminarayanan, B.}, \bibinfo{year}{2019}.
\newblock \bibinfo{title}{Normalizing flows for probabilistic modeling and inference}.
\newblock \bibinfo{journal}{J. Mach. Learn. Res.} \bibinfo{volume}{22}, \bibinfo{pages}{57:1--57:64}.
\bibitem[{Park et~al.(2023)Park, Ryu, Yang, Cho, Kim and Yoon}]{FRM}
\bibinfo{author}{Park, D.H.}, \bibinfo{author}{Ryu, H.}, \bibinfo{author}{Yang, Y.}, \bibinfo{author}{Cho, J.}, \bibinfo{author}{Kim, J.}, \bibinfo{author}{Yoon, K.J.}, \bibinfo{year}{2023}.
\newblock \bibinfo{title}{Leveraging future relationship reasoning for vehicle trajectory prediction}.
\newblock \bibinfo{journal}{ArXiv:2305.14715} .
\bibitem[{Shi et~al.(2022a)Shi, Jiang, Dai and Schiele}]{MTR}
\bibinfo{author}{Shi, S.}, \bibinfo{author}{Jiang, L.}, \bibinfo{author}{Dai, D.}, \bibinfo{author}{Schiele, B.}, \bibinfo{year}{2022}a.
\newblock \bibinfo{title}{Motion transformer with global intention localization and local movement refinement}, in: \bibinfo{booktitle}{Adv. Neural Inf. Process. Syst.}, \bibinfo{publisher}{Curran Associates, Inc.}. pp. \bibinfo{pages}{6531--6543}.
\bibitem[{Shi et~al.(2022b)Shi, Jiang, Dai et~al.}]{MTR-A}
\bibinfo{author}{Shi, S.}, \bibinfo{author}{Jiang, L.}, \bibinfo{author}{Dai, D.}, et~al., \bibinfo{year}{2022}b.
\newblock \bibinfo{title}{Mtr-a: 1st place solution for 2022 waymo open dataset challenge - motion prediction}.
\newblock \bibinfo{journal}{ArXiv:2209.10033} .
\bibitem[{Sohn et~al.(2015)Sohn, Lee and Yan}]{CVAE}
\bibinfo{author}{Sohn, K.}, \bibinfo{author}{Lee, H.}, \bibinfo{author}{Yan, X.}, \bibinfo{year}{2015}.
\newblock \bibinfo{title}{Learning structured output representation using deep conditional generative models}, in: \bibinfo{editor}{Cortes, C.}, \bibinfo{editor}{Lawrence, N.}, \bibinfo{editor}{Lee, D.}, \bibinfo{editor}{Sugiyama, M.}, \bibinfo{editor}{Garnett, R.} (Eds.), \bibinfo{booktitle}{Adv. Neural Inf. Process. Syst.}, \bibinfo{publisher}{Curran Associates, Inc.}
\bibitem[{Varadarajan et~al.(2022)Varadarajan, Hefny, Srivastava, Refaat, Nayakanti, Cornman, Chen, Douillard, Lam, Anguelov and Sapp}]{MultiPath++}
\bibinfo{author}{Varadarajan, B.}, \bibinfo{author}{Hefny, A.}, \bibinfo{author}{Srivastava, A.}, \bibinfo{author}{Refaat, K.S.}, \bibinfo{author}{Nayakanti, N.}, \bibinfo{author}{Cornman, A.}, \bibinfo{author}{Chen, K.}, \bibinfo{author}{Douillard, B.}, \bibinfo{author}{Lam, C.P.}, \bibinfo{author}{Anguelov, D.}, \bibinfo{author}{Sapp, B.}, \bibinfo{year}{2022}.
\newblock \bibinfo{title}{Multipath++: Efficient information fusion and trajectory aggregation for behavior prediction}, in: \bibinfo{booktitle}{IEEE Int. Conf. Robot. Autom.}, pp. \bibinfo{pages}{7814--7821}.
\newblock \DOIprefix\doi{10.1109/ICRA46639.2022.9812107}.
\bibitem[{Velickovic et~al.(2017)Velickovic, Cucurull, Casanova, Romero, Lio’ and Bengio}]{GAT}
\bibinfo{author}{Velickovic, P.}, \bibinfo{author}{Cucurull, G.}, \bibinfo{author}{Casanova, A.}, \bibinfo{author}{Romero, A.}, \bibinfo{author}{Lio’, P.}, \bibinfo{author}{Bengio, Y.}, \bibinfo{year}{2017}.
\newblock \bibinfo{title}{Graph attention networks}.
\newblock \bibinfo{journal}{ArXiv:1710.10903} .
\bibitem[{Wang et~al.(2022)Wang, Ye, Gu and Chen}]{LTP}
\bibinfo{author}{Wang, J.}, \bibinfo{author}{Ye, T.}, \bibinfo{author}{Gu, Z.}, \bibinfo{author}{Chen, J.}, \bibinfo{year}{2022}.
\newblock \bibinfo{title}{Ltp: Lane-based trajectory prediction for autonomous driving}, in: \bibinfo{booktitle}{Proc. IEEE/CVF Conf. Comput. Vis. Pattern Recognit.}, pp. \bibinfo{pages}{17113--17121}.
\newblock \DOIprefix\doi{10.1109/CVPR52688.2022.01662}.
\bibitem[{Wang et~al.(2023)Wang, Su, Da and Yang}]{prophnet}
\bibinfo{author}{Wang, X.}, \bibinfo{author}{Su, T.}, \bibinfo{author}{Da, F.}, \bibinfo{author}{Yang, X.}, \bibinfo{year}{2023}.
\newblock \bibinfo{title}{Prophnet: Efficient agent-centric motion forecasting with anchor-informed proposals}, in: \bibinfo{booktitle}{Proc. IEEE/CVF Conf. Comput. Vis. Pattern Recognit.}, pp. \bibinfo{pages}{21995--22003}.
\newblock \DOIprefix\doi{10.1109/CVPR52729.2023.02106}.
\bibitem[{Wilson et~al.(2023)Wilson, Qi, Agarwal, Lambert, Singh, Khandelwal, Pan, Kumar, Hartnett, Pontes, Ramanan, Carr and Hays}]{argoverse2}
\bibinfo{author}{Wilson, B.}, \bibinfo{author}{Qi, W.}, \bibinfo{author}{Agarwal, T.}, \bibinfo{author}{Lambert, J.}, \bibinfo{author}{Singh, J.}, \bibinfo{author}{Khandelwal, S.}, \bibinfo{author}{Pan, B.}, \bibinfo{author}{Kumar, R.}, \bibinfo{author}{Hartnett, A.}, \bibinfo{author}{Pontes, J.K.}, \bibinfo{author}{Ramanan, D.}, \bibinfo{author}{Carr, P.}, \bibinfo{author}{Hays, J.}, \bibinfo{year}{2023}.
\newblock \bibinfo{title}{Argoverse 2: Next generation datasets for self-driving perception and forecasting}.
\newblock \bibinfo{journal}{ArXiv:2301.00493} \href{http://arxiv.org/abs/2301.00493}{\tt arXiv:2301.00493}.
\bibitem[{Xu et~al.(2022a)Xu, Li, Ni, Zhang and Chen}]{groupnet}
\bibinfo{author}{Xu, C.}, \bibinfo{author}{Li, M.}, \bibinfo{author}{Ni, Z.}, \bibinfo{author}{Zhang, Y.}, \bibinfo{author}{Chen, S.}, \bibinfo{year}{2022}a.
\newblock \bibinfo{title}{Groupnet: Multiscale hypergraph neural networks for trajectory prediction with relational reasoning}, in: \bibinfo{booktitle}{Proc. IEEE/CVF Conf. Comput. Vis. Pattern Recognit.}, pp. \bibinfo{pages}{6488--6497}.
\newblock \DOIprefix\doi{10.1109/CVPR52688.2022.00639}.
\bibitem[{Xu et~al.(2022b)Xu, Wei, Tang, Yin, Zhang and Chen}]{Dygroupnet}
\bibinfo{author}{Xu, C.}, \bibinfo{author}{Wei, Y.}, \bibinfo{author}{Tang, B.}, \bibinfo{author}{Yin, S.}, \bibinfo{author}{Zhang, Y.}, \bibinfo{author}{Chen, S.}, \bibinfo{year}{2022}b.
\newblock \bibinfo{title}{Dynamic-group-aware networks for multi-agent trajectory prediction with relational reasoning}.
\newblock \bibinfo{journal}{ArXiv:2206.13114} \href{http://arxiv.org/abs/2206.13114}{\tt arXiv:2206.13114}.
\bibitem[{Xu et~al.(2022c)Xu, Hayet and Karamouzas}]{socialvae2022}
\bibinfo{author}{Xu, P.}, \bibinfo{author}{Hayet, J.B.}, \bibinfo{author}{Karamouzas, I.}, \bibinfo{year}{2022}c.
\newblock \bibinfo{title}{Socialvae: Human trajectory prediction using timewise latents}, in: \bibinfo{booktitle}{Proc. IEEE/CVF Eur. Conf. Comput. Vis.}, \bibinfo{organization}{Springer}. pp. \bibinfo{pages}{511--528}.
\newblock \DOIprefix\doi{10.1007/978-3-031-19772-7_30}.
\bibitem[{Ye et~al.(2022)Ye, Wang, Chen, Wang, Wu and Lu}]{Ye2022}
\bibinfo{author}{Ye, L.}, \bibinfo{author}{Wang, Z.}, \bibinfo{author}{Chen, X.}, \bibinfo{author}{Wang, J.}, \bibinfo{author}{Wu, K.}, \bibinfo{author}{Lu, K.}, \bibinfo{year}{2022}.
\newblock \bibinfo{title}{Gsan: Graph self-attention network for learning spatial–temporal interaction representation in autonomous driving}.
\newblock \bibinfo{journal}{IEEE Internet Things J.} \bibinfo{volume}{9}, \bibinfo{pages}{9190--9204}.
\bibitem[{Ye et~al.(2021)Ye, Cao and Chen}]{TPCN}
\bibinfo{author}{Ye, M.}, \bibinfo{author}{Cao, T.}, \bibinfo{author}{Chen, Q.}, \bibinfo{year}{2021}.
\newblock \bibinfo{title}{Tpcn: Temporal point cloud networks for motion forecasting}, in: \bibinfo{booktitle}{Proc. IEEE/CVF Conf. Comput. Vis. Pattern Recognit.}, pp. \bibinfo{pages}{11313--11322}.
\newblock \DOIprefix\doi{10.1109/CVPR46437.2021.01116}.
\bibitem[{Zeng et~al.(2021)Zeng, Liang, Liao and Urtasun}]{Lanercnn}
\bibinfo{author}{Zeng, W.}, \bibinfo{author}{Liang, M.}, \bibinfo{author}{Liao, R.}, \bibinfo{author}{Urtasun, R.}, \bibinfo{year}{2021}.
\newblock \bibinfo{title}{Lanercnn: Distributed representations for graph-centric motion forecasting}, in: \bibinfo{booktitle}{Proc. IEEE/RSJ Int. Conf. Intell. Rob. Syst.}, pp. \bibinfo{pages}{532--539}.
\bibitem[{Zhao et~al.(2021)Zhao, Gao, Lan, Sun, Sapp, Varadarajan, Shen, Shen, Chai, Schmid, Li and Anguelov}]{TNT}
\bibinfo{author}{Zhao, H.}, \bibinfo{author}{Gao, J.}, \bibinfo{author}{Lan, T.}, \bibinfo{author}{Sun, C.}, \bibinfo{author}{Sapp, B.}, \bibinfo{author}{Varadarajan, B.}, \bibinfo{author}{Shen, Y.}, \bibinfo{author}{Shen, Y.}, \bibinfo{author}{Chai, Y.}, \bibinfo{author}{Schmid, C.}, \bibinfo{author}{Li, C.}, \bibinfo{author}{Anguelov, D.}, \bibinfo{year}{2021}.
\newblock \bibinfo{title}{Tnt: Target-driven trajectory prediction}, in: \bibinfo{booktitle}{Proc. Conf. Robot Learn.}, \bibinfo{publisher}{PMLR}. pp. \bibinfo{pages}{895--904}.
\bibitem[{Zhou et~al.(2022)Zhou, Ye, Wang, Wu and Lu}]{HiVT}
\bibinfo{author}{Zhou, Z.}, \bibinfo{author}{Ye, L.}, \bibinfo{author}{Wang, J.}, \bibinfo{author}{Wu, K.}, \bibinfo{author}{Lu, K.}, \bibinfo{year}{2022}.
\newblock \bibinfo{title}{Hivt: Hierarchical vector transformer for multi-agent motion prediction}, in: \bibinfo{booktitle}{Proc. IEEE/CVF Conf. Comput. Vis. Pattern Recognit.}, pp. \bibinfo{pages}{8813--8823}.
\newblock \DOIprefix\doi{10.1109/CVPR52688.2022.00862}.

\end{thebibliography}






\end{document}